\documentclass{article}

\usepackage{microtype}
\usepackage{graphicx}
\usepackage{subcaption}
\usepackage{booktabs} % for professional tables
\usepackage{dblfloatfix}

\usepackage{hyperref}

\usepackage[preprint]{icml2026}

\usepackage{amsmath}
\usepackage{amssymb}
\usepackage{mathtools}
\usepackage{amsthm}

\usepackage[capitalize,noabbrev]{cleveref}

\theoremstyle{plain}

\theoremstyle{definition}

\theoremstyle{remark}

\usepackage[textsize=tiny]{todonotes}

\icmltitlerunning{Solving adversarial examples requires solving exponential misalignment}

\begin{document}

\twocolumn[
  \icmltitle{Solving adversarial examples requires solving exponential misalignment}



  \icmlsetsymbol{equal}{*}

  \begin{icmlauthorlist}
    \icmlauthor{Alessandro Salvatore}{1}
    \icmlauthor{Stanislav Fort}{2}
    \icmlauthor{Surya Ganguli}{1}
\end{icmlauthorlist}

  \icmlaffiliation{1}{Department of Applied Physics, Stanford University, CA, USA}
  \icmlaffiliation{2}{Aisle, San Francisco, CA, USA}
  
  \icmlcorrespondingauthor{Alessandro Salvatore}{alesaso@stanford.edu}
  \icmlcorrespondingauthor{Surya Ganguli}{sganguli@stanford.edu}

  \icmlkeywords{Machine Learning, ICML}

  \vskip 0.3in
]



\printAffiliationsAndNotice{}  % no special notice (required even if empty)

\begin{abstract}

Adversarial attacks—input perturbations imperceptible to humans that fool neural networks—remain both a persistent failure mode in machine learning, and a phenomenon with mysterious origins.  
To shed light, we define and analyze a network's perceptual manifold (PM) for a class concept as the space of all inputs confidently assigned to that class by the network. 
We find, strikingly, that the dimensionalities of neural network PMs are orders of magnitude {\it higher} than those of natural human concepts. 
%For example, PMs of ResNets on CIFAR10 occupy {\it 3060 dimensions} out 3072 for CIFAR10 images, while PMs of CLIP trained ViTs occupy {\it 135,000 dimensions} out of 150,000 for ImageNet scale images. In contrast, natural image concepts are {\it only 20} dimensions.  
Since volume typically grows exponentially with dimension, this suggests {\it exponential misalignment} between machines and humans, with exponentially many inputs confidently assigned to concepts by machines but not humans.  
Furthermore, this provides a natural geometric hypothesis for the origin of adversarial examples: because a network's PM fills such a large region of input space, {\it any} input will be very close to {\it any} class concept's PM.  
Our hypothesis thus suggests that adversarial robustness cannot be attained without dimensional alignment of machine and human PMs, and therefore makes strong predictions: both robust accuracy and distance to any PM should be negatively correlated with the PM dimension. 
We confirmed these predictions across 18 different networks of varying robust accuracy. Crucially, we find even the most robust networks are still exponentially misaligned, and only the few PMs whose dimensionality approaches that of human concepts exhibit alignment to human perception. 
Our results connect the fields of alignment and adversarial examples, and suggest the curse of high dimensionality of machine PMs is a major impediment to adversarial robustness.

\end{abstract}

\begin{figure}[t!]
    \includegraphics[width=1\linewidth]{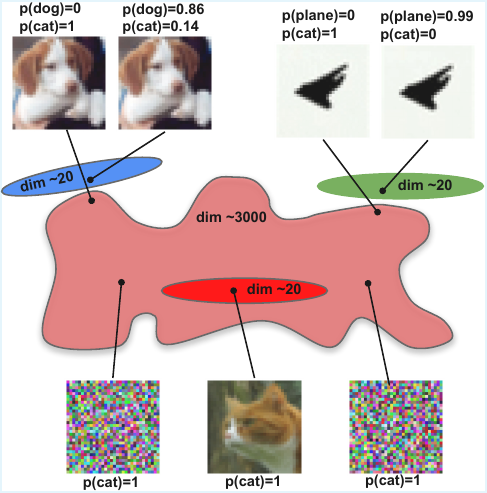}
    \caption{
    Visualization of our main argument: We show that a network's perceptual manifold (PM) for any class concept (e.g. cat), defined to be the set of {\it all} images confidently perceived by the network as that class, is {\it extremely high dimensional}: $3060$ out of a total of $3072$ for CIFAR10 (large red manifold) and $\approx 135,000$ out of $150,528$ for CLIP and ImageNet. In contrast natural images perceived by humans as any class (e.g. dogs, airplanes or cats (blue, green and bright red manifolds)) are only $\approx 20$ dimensional.  This indicates that machine and human PMs for any concept are {\it exponentially misaligned}: there are exponentially many inputs confidently perceived as any given concept by machines, but not by humans (e.g. the two noise images in the network's cat perceptual manifold). This exponential misalignment also explains the origin of adversarial examples: e.g. because the network's cat PM fills up so much of image space, any other input (e.g. dog or airplane) is extremely close to it.
    }   
    \label{fig:schematic}
\end{figure}

\section{Introduction}

Given the advent of remarkable and exceedingly powerful AI systems, the task of solving AI alignment, or successfully aligning the behavior of these powerful AI systems to human values and intentions, has recently emerged as a fundamental and important area of research (see e.g. \citep{ji_ai_2025-1, ji_ai_2025} for reviews).  

At the same time, the curious case of adversarial attacks, or small input perturbations that fool neural networks but are essentially imperceptible to humans, were discovered more than a decade ago \citep{szegedy_intriguing_2014}. Despite progress on increasing the adversarial robustness of neural networks, as documented in the RobustBench leaderboard \citep{croce_robustbench_2021}, even after a decade of work, neural networks are still nowhere near as robust as human vision, or at least the perturbations that fool neural networks do {\it not} fool humans.  Thus the inability to achieve robust classification has remained a major failure mode in machine learning. Moreover, the geometric reasons for the persistence of adversarial examples and our inability to remove them remain mysterious.  Why is it the case that in standard neural networks, {\it any} image is close to that of {\it any other} class, and why has it been so hard to remove this property?   

Here we shed light on these questions by connecting adversarial examples to the notion of (mis)alignment. In particular, we show that adversarial examples are due to an exceedingly severe form of {\it exponential misalignment} between the nature of machine versus human perception.  The core of our argument is visualized and explained in Fig.~\ref{fig:schematic}. We show this misalignment by showing that the dimensionality of the space of images confidently perceived by a machine as belonging to a particular class concept (the machine's {\it perceptual manifold} (PM) for that concept) is {\it orders of magnitude higher} than that of natural images perceived by humans as belonging to that concept.  Thus the {\it curse of high dimensionality} of machine perceptual manifolds rears its ugly head as the main geometric reason for the existence of adversarial examples, as well as the main difficulty in removing them. Our work suggests that adversarial examples cannot be solved unless we can train networks to have perceptual manifolds that are as low dimensional as that of humans: i.e. we must at the very least {\it dimensionally align} machine and human perception.

This motivates a renewed interest from the perspective of modern alignment in solving the old problem of adversarial examples: {\it they are a prototypical warmup problem of misalignment at the level of perception}.  Our work also serves as a cautionary tale for the more general field of alignment: achieving desirable AI behavior aligned with human intentions and values over an exponentially large space of image and text inputs may be even harder than solving the decade long unsolved problem of adversarial examples, which similarly involves achieving desirable perceptual behavior over an exponentially large space of sensory inputs.  

\paragraph{Related work.} A rich body of literature attempts to explain the existence of adversarial examples, proposing mechanisms such as local linearity \citep{goodfellow_explaining_2015}, predictive but non-robust feature reliance \citep{ilyas_adversarial_2019}, non-robust classification boundary \citep{tanay_boundary_2016}, high-dimensional geometry \citep{gilmer_adversarial_2018}, and more. We relegate a full discussion of these background theories and their relation to ours to App.~\ref{sec: Background and Related Work}. 

\paragraph{Outline.} In Sec.~\ref{sec:overall-framework} we describe our overall framework, including defining perceptual manifolds, and describing how to sample from them and measure their dimensionality. In Sec.~\ref{sec:fragility} and Sec.~\ref{sec:clip} we demonstrate severe exponential misalignment in ResNets on CIFAR10, CLIP models on CIFAR10 and LSUN, and CNNs on ImageNet. In Sec.~\ref{sec:fragility} we also provide a theoretical argument connecting exponential misalignment to adversarial fragility in a toy model of perceptual manifolds (PMs). In Sec.~\ref{sec:correlation} we test two key predictions of our theory in 18 different networks of varying degrees of robust accuracy: both robust accuracy and distance of any image to the PM {\it increase} as PM dimensionality {\it decreases}. This provides further evidence that reducing exponential misalignment may be a necessary prerequisite for achieving adversarial robustness. In Sec.~\ref{sec: persistent misalignment} we further explore (mis)alignment between machine and human perception at the level of PM eigenstructure and individual samples from PMs.  Crucially we find even highly robust networks are still exponentially misaligned. However we see sparks of alignment in the most robust networks {\it only} for classes with the {\it lowest} dimensional PMs approaching that of natural human PM dimensions. This again highlights the fundamental importance of dimensional alignment between machine and human PMs as a likely prerequisite for solving adversarial examples. Finally, in Sec.~\ref{sec: imagenet}, we extend our analysis to ImageNet-1K \citep{russakovsky_imagenet_2015}, where the observed trends mirror the results reported in the main text.

.

\section{Overall framework}
\label{sec:overall-framework}

 Here we define a network's perceptual manifold for a class concept (Sec.~\ref{subsec:PMdef}), explain how to sample from this manifold (Sec.~\ref{subsec:PMsam}), and then describe two measures of its dimensionality (Sec.~\ref{subsec:PMdim}). Finally, we review the notions of adversarial attacks and robust accuracy (Sec.~\ref{subsec:robacc}).  In the following sections we will connect the disparate notions of robust accuracy and perceptual manifold dimensionality.    

\subsection{A network's perceptual manifold for a class concept}
\label{subsec:PMdef}

Let $p(c \mid x)$ denote the probability that a neural network classifier assigns to a class $c$ given input $x$. In this work, we focus on $D$ dimensional image inputs with input pixel values normalized between $0$ and $1$, so $x \in [0,1]^D$. We define the network's perceptual manifold (PM) for any one of its possible output class concepts $c$ to be the set of {\it all} inputs $x$ that the network assigns to class $c$ with high confidence: 
\begin{equation}
    \text{PM} \equiv \{ x \in [0,1]^D \mid p(c \mid x) > p_0 \},
    \label{eq: PM definition}
\end{equation}
where $p_0$ is the threshold confidence level. Unless otherwise stated, we will choose a confidence level of $p_0 = 0.9$, which is much higher than that of robust networks ($60\%$ - $70\%$ on correctly classified CIFAR10 \citep{krizhevsky_learning_nodate} test images) and comparable to that of standard non-robust networks (often $\geq 95\%$).
Crucially, the perceptual manifold characterizes the subset of the input space that the network \textit{confidently} perceives as exemplars of concept $c$. 
%This definition differs fundamentally from the standard decision region, which consists of all inputs where $c$ is simply the most probable class (i.e., $x$ such that $c = \operatorname*{argmax}_{c'} p(c' \mid x)$). While the decision region exhaustively partitions the input space, forcing the model to assign a label even to ambiguous inputs or noise, the threshold $p_0$ effectively creates a rejection class. 
A main goal is to explore the misalignment between this machine PM and the human PM of natural images that humans would confidently classify as exemplars of concept $c$.

\subsection{Sampling a network's perceptual manifold}
\label{subsec:PMsam}
To explore the contents of a network's PM for a class $c$, we perform random sampling of the manifold via projected gradient ascent (PGA), as follows.  Each sample is obtained by starting from a random initial noise image $x_0 \sim \mathcal{U}[0,1]^D$ drawn uniformly from the $D$-dimensional unit hypercube. We then perform multiple iterations of a $2$ step process of: (1) gradient ascent on the class log-probability $\ln p(c|x)$, and (2) projection onto the image hypercube $[0,1]^D$.  We continue these alternating steps until we reach an iteration $t$ with image $x_t \in [0,1]^D $ satisfying $p(c \mid x) > p_0$. See App.~\ref{sec: PMsam appendix} for pseudo-code and details.

\subsection{Dimensionality of perceptual manifolds}
\label{subsec:PMdim}

We consider two measures of dimensionality of either natural image manifolds, or machine perceptual manifolds. Each measure of dimensionality can be estimated from a finite set of samples, described as follows.

\paragraph{Participation ratio (PR) dimensionality.} The PR measures the effective dimensionality based on the eigenvalues of the sample covariance matrix. For a set of $N$ samples $\{\mathbf{x}_i\}_{i=1}^N$ in $\mathbb{R}^D$ with covariance matrix $\mathbf{C} = \frac{1}{N}\sum_{i=1}^N (\mathbf{x}_i - \bar{\mathbf{x}})(\mathbf{x}_i - \bar{\mathbf{x}})^\top$, the PR dimension is defined as
\begin{equation}
d_{\text{PR}} = \frac{\left(\sum_{j=1}^D \lambda_j\right)^2}{\sum_{j=1}^D \lambda_j^2} = \frac{\operatorname{tr}(\mathbf{C})^2}{\operatorname{tr}(\mathbf{C}^2)},
\end{equation}
where $\{\lambda_j\}_{j=1}^D$ are the eigenvalues of $\mathbf{C}$. This metric quantifies the number of significant directions of variance, yielding a value between $1$ (when only $1$ eigenvalue is nonzero) and $D$ (when all eigenvalues are identical). 

\paragraph{Two nearest neighbor (2NN) dimensionality.} 

The 2 nearest neighbor (2NN) method \citep{facco_estimating_2017} estimates the \textit{intrinsic dimension} of a manifold, i.e. the minimum number of degrees of freedom required to traverse the manifold, from a set of $N$ samples. 
See App.~\ref{sec:estindim} for the  precise procedure we use to estimate the 2NN dimension, which we denote by $d_{\text{2NN}}$. 
We note that if data is sampled from a manifold of intrinsic dimension $d$, unless the number of samples $N$ is exponential in $d$, the 2NN estimate $d_{\text{2NN}}$ will be an underestimate of the true $d$.  
Consequently, our reported values of $d_{\text{2NN}}$ should be interpreted as lower bounds on the true intrinsic dimension.  
Nevertheless we will find that our estimated lower bounds on the intrinsic dimension of machine PMs still far exceed that of natural image concepts.

\subsection{Review of adversarial attacks and robust accuracy}
\label{subsec:robacc}

A fundamental goal of our work is to connect the high dimensionality of machine PMs to the lack of robust accuracy under adversarial attacks. We therefore briefly review the notion of adversarial attacks and robust accuracy.  

Given an input $x$ with correct class label $c^*$, an adversarial attack seeks an input perturbation $\delta$ obeying two conditions:
\begin{equation}
\label{eq:advattack}
\arg\max_{c'} p(c' | x + \delta) \neq c^* \quad \text{and} \quad \|\delta\|_p \le \epsilon.
\end{equation}
The first condition says the perturbed input $x+\delta$ should fool the network into classifying $x+\delta$ as a {\it different} class {\it not} equal to the correct class $c^*$ of $x$.  The second condition says the perturbation should not exceed a small radius $\epsilon$ in $L_p$ norm, often rendering the perturbation imperceptible to humans.  In this work, we focus on the most common CIFAR case of $L_\infty$ norm with radius of $\epsilon = 8/255$.

The robust accuracy of a network is then defined as the fraction of test images $x$ for which {\it no} perturbation $\delta$ exists satisfying the two conditions in \eqref{eq:advattack}.  In essence a network correctly and robustly classifies an input $x$ if it outputs the correct class label $c^*$ not only for $x$ but for {\it all} images $x+\delta$ within the $L_p$ ball of radius $\epsilon$ centered at $x$. In practice of course, one cannot exhaustively search the entire $L_p$ ball.  One instead has to resort to some ensemble of strong attack algorithms and we say an input is robustly classified if the attack algorithms fail to find a perturbation $\delta$ satisfying \eqref{eq:advattack}. The failure probability of such attacks thus constitutes an estimate of the robust accuracy of a network.  This estimate is actually an upper bound on the true robust accuracy, as there may exist as yet undiscovered more powerful attacks with lower failure probability.  In our work we employ the AutoAttack ensemble \citep{croce_reliable_2020} which is used to score the robust accuracy of networks in the RobustBench leaderboard \citep{croce_robustbench_2021}.

\section{Fragility from exponential misalignment}
\label{sec:fragility}
\begin{figure*}[t]
    \centering
    \includegraphics[width=0.9\linewidth]{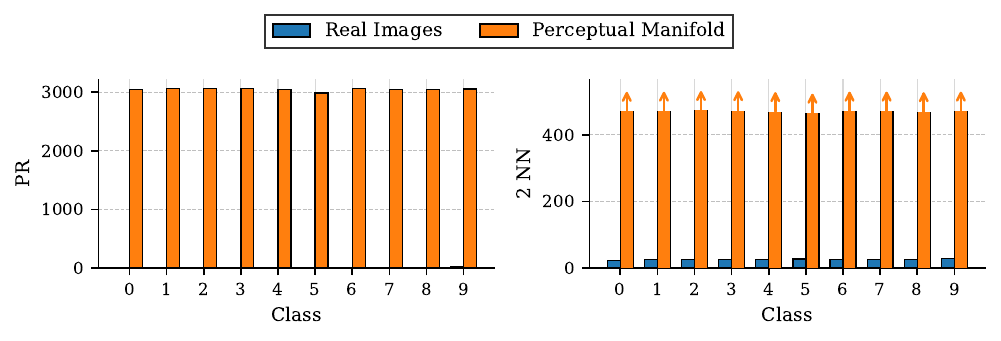}
    \caption{Comparison between the dimensionality (Participation Ratio on the left and Two Nearest Neighbors on the right) of the Perceptual Manifold of a WideResNet-28-10 (clean accuracy of $94.78\%$ and robust accuracy of $0\%$) to that of the natural images for each class. In the left plot, the PR of the natural images is $\approx 10$, which makes it barely visible. The arrows in the right plot indicate that those values are lower bounds. The excessive dimensionality of machine PMs relative to their natural counterparts signals exponential misalignment.}
    \label{fig: barplot exponential misalignment}
\end{figure*}

In Sec.~\ref{subsec: standard is exponentially misaligned} we first show that non-robust networks that are highly fragile to adversarial attacks do indeed exhibit exponential misalignment. In Sec.~\ref{subsec:georel} we then demonstrate, in a toy model, a causal relationship between exponential misalignment and fragility to adversarial attacks that relies on the geometry of distances between points and simple high dimensional manifolds. These results suggest the hypothesis that adversarial robustness with high robust accuracy can only be achieved if exponential misalignment can be solved. We provide further evidence for this hypothesis in Sec.~\ref{sec:correlation}.  

\subsection{Non-robust networks are exponentially misaligned}
\label{subsec: standard is exponentially misaligned}

To show that standard non-robust networks are exponentially misaligned, in Fig.~\ref{fig: barplot exponential misalignment} we compare the dimensionality of the set of natural images to that of the PM of a standard non-robust model for each class of CIFAR-10. The latter exceeds the former by orders of magnitude: the Participation Ratio (PR) and Two Nearest Neighbors (2NN) estimates for natural images are $\approx 10$ and $\approx 20$, while those of the model's PM are $\approx 3060$ and $\geq 500$ respectively (noting that the latter is a lower bound).  Again, since volume typically grows exponentially with dimension, this orders of magnitude dimensional discrepancy between machine and natural image PMs indicates exponential misalignment: there are {\it exponentially} many images artificial networks confidently assign to any concept that humans do not (see e.g. Fig.~\ref{fig: sparks of alignment}). We repeat this analysis on ImageNet-1K in \cref{subsec: imagenet exponential misalignment} with consistent results. Crucially, this catastrophic dimensional gap is not an artifact of low-resolution data; on ImageNet-1K, the PM of a standard ResNet-50 occupies over 130,000 out of 150,528 dimensions, while the natural image concepts remain roughly 20-dimensional.
%The unusual inequality $PR < 2NN$ for real images stems from the dataset's extreme spectral anisotropy: the Participation Ratio is heavily biased toward the few high-variance modes that dominate global statistics, whereas the 2NN estimator detects additional 'hidden' dimensions effectively because their local spread, though negligible compared to the principal components, exceeds the typical nearest-neighbor distance required for resolvability.

\subsection{Geometry relating misalignment to fragility}
\label{subsec:georel}

\begin{figure}
    \centering
    \includegraphics[width=0.9\linewidth]{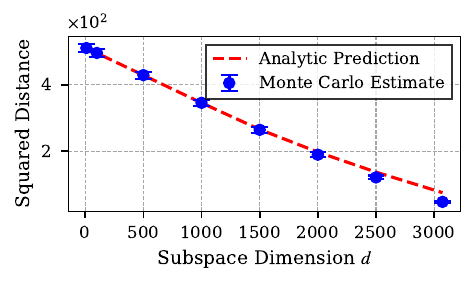}
    \caption{Monte Carlo estimate of the squared distance between a random point sampled from unit hypercube and the boundary of a $d$-dimensional ellipsoid. The principal axes are sampled from a uniform distribution $\mathcal{U}[6,30]$, so the volume of the full 3072 dimensional ellipsoid is roughly equal to the estimated volume of the non robust model's PM (when approximating it by an ellipsoid). Error bars cover the $\pm 1\sigma$ interval. Theory is the red dashed curve.}
    \label{fig:toy model}
\end{figure}

We now provide a logical connection between a networks PM dimensionality and the network's fragility to bounded norm adversarial attacks.  Intuitively, the higher the dimension of a manifold, the more of ambient space it will fill up, and the closer {\it any} point in the ambient space will be to that manifold. In the context of CIFAR10 images, if the network's PM is $3060$ dimensional in an ambient space of $3072$ dimensions, it fills up nearly all of image space, and so {\it any} image $x\in [0,1]^D$ not in the PM, should be very close to the PM, and therefore is highly likely to admit a successful bounded norm adversarial attack.  

While it is not possible to theoretically predict the expected distance between a random point and a manifold, knowing only the dimensionality of the manifold, we can make the above intuition quantitatively precise in a toy model of ellipsoidal PMs as follows. 
Let the PM be a $d$-dimensional ellipsoid in a $D$-dimensional ambient space, $\mathcal{E} = \left\{ x \mid x = Vy + c, \quad y^\top H y \leq 1 \right\}$, centered at $c\sim \mathcal{U}[0,1]^D$, where $V\in\mathbb{R}^{D\times d}$, $x\in\mathbb{R}^D$ and $y\in\mathbb{R}^d$. In App. \ref{sec:toy_model_distance}, we derive the expected squared distance between a random point $x\sim\mathcal{U}[0,1]^D$ and the ellipsoid:
\begin{equation}
    \mathbb{E}\left[\|x - \partial\mathcal{E}\|^2 \right] \approx \frac{D - d}{6} + \max\left\{0,\sqrt{\frac{d}{6}} - R_{eff}\right\}^2, 
\end{equation}
where $R_{eff}$ is the root mean square of principal radii of the ellipsoid (see App.~\ref{sec:toy_model_distance}).
%(i.e., in terms of the eigenvalues $\{\lambda_i\}$ of $H$: $R_{eff} = \left(\prod_i^d\frac{1}{\sqrt{\lambda_i}}\right)^{\frac{1}{d}})$. 

This analytical result highlights a linear decrease in expected distance to the PM as its dimensionality $d$ increases, as verified numerically in Fig.~\ref{fig:toy model}. This dependence leads to a massive change: for low-dimensional ellipsoids, the squared distance in the $3072$ dimensional ambient space of CIFAR10 is $\approx 500$, whereas for full-dimensional manifolds, the distance collapses to $\approx 50$.  For a distance scale reference, two uniformly random points in a $D$ dimensional unit hypercube have expected squared distance $\frac{D}{6} = 512$ for $D=3072$.  Thus a point is exceedingly close to a random high dimensional manifold relative to any other random point.  This connects the high dimensionality of network PMs to adversarial fragility of networks in a toy setting, but the connection is more general, as we confirm in Sec.~\ref{sec:correlation}.

\begin{figure*}
    \centering
    \includegraphics[width=0.9\linewidth]{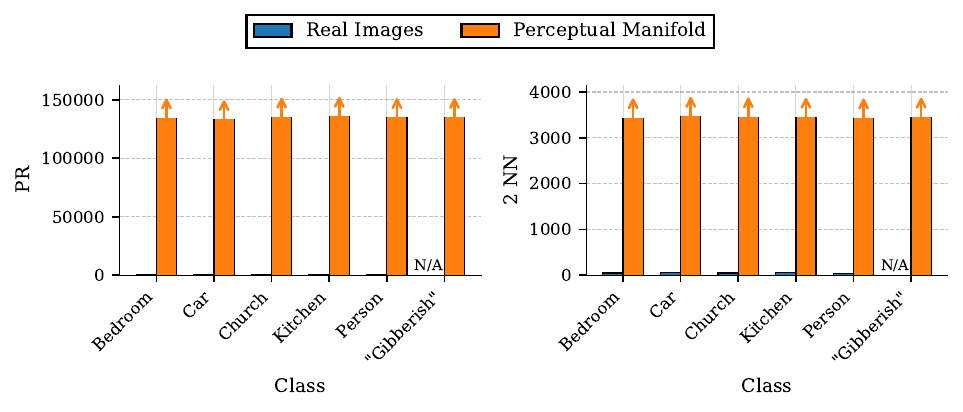}
    \caption{Dimensionality comparison (Participation Ratio and Two Nearest Neighbors) of the CLIP Perceptual Manifold versus natural images (LSUN dataset). Arrows indicate that those reported are lower bounds on the actual value, we include a plot of how the predicted PR and 2NN scale with dataset size in \cref{fig: scaling clip pr} and \cref{fig: scaling clip two nn}. We include a ``Gibberish'' control prompt: ``kjdbfkw hsafj asjf gjkbg''; note that the natural dimensionality of the gibberish class is undefined (N/A) as no natural images correspond to it.}
    \label{fig: exponential misalignment clip}
\end{figure*}

\section{Exponential misalignment in CLIP}\label{sec:clip}

We now investigate whether exponential misalignment is an artifact of standard supervised training on fixed class labels, or if it persists in foundation models trained on massive datasets via different objectives. Specifically, we analyze CLIP \citep{radford_learning_2021}, a model trained via contrastive learning to align image and text representations. 

\subsection{CLIP as a zero shot classifier} 
CLIP consists of an image encoder $f(\cdot)$ and a text encoder $g(\cdot)$ that map inputs into a shared embedding space. When used as a zero-shot classifier, one typically computes the cosine similarity between an image embedding and the embeddings of a set of class descriptions (e.g. ``a photo of a {class}''), assigning the image to the class with the highest similarity. Despite its impressive generalization, CLIP has been shown to be vulnerable to adversarial examples \citep{fort_adversarial_2021}, suggesting that its perceptual manifolds may share the pathological dimensionality of standard classifiers.

\subsection{Defining and Sampling CLIP's Perceptual Manifold}

We define the CLIP Perceptual Manifold (PM) for a target text $t$ as the level set of inputs $x$ that satisfy a cosine similarity lower bound $c_0$:\begin{equation}\text{PM}_{\text{CLIP}}(t) = \left\{ x \in [0,1]^D \middle| \frac{f(x) \cdot g(t)}{\|f(x)\|\|g(t)\|} \geq c_0 \right\}.\end{equation}
We employ a strict threshold of $c_0 = 0.6$. This is a good high threshold, as natural images corresponding to vague prompts (e.g., ``\textit{an image of a cat}'') typically achieve similarity scores of only $\approx 0.3$, and white noise images yield a baseline similarity of $\approx 0.25$ against these same prompts.\footnote{This noise baseline drops to $\approx 0.17$ for highly specific, lengthy prompts (e.g., detailed photorealistic descriptions).} Consequently, to ensure semantic specificity and isolate the target concept from high-similarity background noise, we require a threshold significantly exceeding the natural distribution. To sample from this high-confidence PM, we utilize Projected Gradient Ascent (PGA) as in Sec.~\ref{subsec:PMsam}.

\subsection{CIFAR-like experiment}
\begin{table}[b!]
    \centering
    \caption{Dimensionality comparison between the Natural Manifold and CLIP's Perceptual Manifold (PM) in the CIFAR-like setting ($32 \times 32$ inputs). The PM is consistently orders of magnitude larger than the natural data manifold.}
    \label{tab:cifar_clip}
    \begin{tabular}{lcccc}
        \toprule
        & \multicolumn{2}{c}{\textbf{Natural Manifold}} & \multicolumn{2}{c}{\textbf{Perceptual Manifold}} \\
        \cmidrule(lr){2-3} \cmidrule(lr){4-5}
        \textbf{Class} & \textbf{PR} & \textbf{2NN} & \textbf{PR} & \textbf{2NN} \\
        \midrule
        Cat & 10 & 21 & 2650 & $\geq$412 \\
        Truck & 11 & 23 & 2652 & $\geq$420 \\
        \bottomrule
    \end{tabular}
\end{table}

To allow for a direct comparison with the CIFAR10 classifiers analyzed in \cref{subsec: standard is exponentially misaligned}, we restrict the optimization to a $32 \times 32 \times 3$ input tensor (which is then upsampled before it is passed to the CLIP image encoder).  We target two specific prompts: ``a photo of a cat'' and ``a photo of a truck'', corresponding to standard CIFAR-10 classes. The results, summarized in \cref{tab:cifar_clip}, reveal that the PM is again about {\it 2 orders of magnitude higher} dimensional than the natural image dimensions, despite the high confidence threshold chosen.  This stark contrast reveals that CLIP exhibits exponential misalignment comparable to that of standard CNNs.

\subsection{Full dimensionality experiments} 

\begin{figure}[b!]
    \centering
    \includegraphics[width=0.9\linewidth]{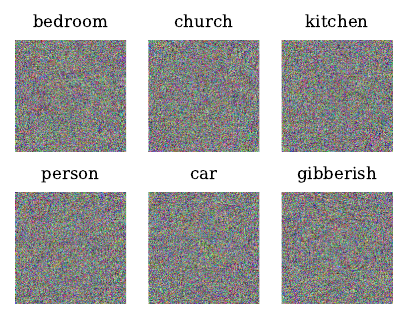}
    \caption{Representative samples from CLIP's Perceptual Manifold for valid descriptions (e.g., ``a photo of a bedroom'') and the control prompt ``kjdbfkw hsafj asjf gjkbg''. Visually, all samples appear as noise, again indicating exponential misalignment between machine and human perception.}
    \label{fig: clip samples}
\end{figure}

We next analyze the manifold in the native input space of a CLIP ViT-B/32 image encoder ($224 \times 224 \times 3$, $D = 150,528$). We select target concepts from the LSUN dataset \citep{yu_lsun_2016} (e.g., ``a photo of a bedroom'') and sample the corresponding PMs. Crucially, to distinguish between dimensionality driven by semantic complexity and structural artifacts, we also include a semantically vacuous control prompt: ``kjdbfkw hsafj asjf gjkbg''. As illustrated in \cref{fig: exponential misalignment clip}, the dimensionalities of CLIP's PMs dwarf that of real images {\it by 4 orders of magnitude} and remain statistically indistinguishable between meaningful concepts and the gibberish control. We measure a PR of $\geq 135,000$ and a 2NN of $\geq 3,450$ for the CLIP PMs, implying they fill the ambient space almost entirely, whereas the natural image manifolds have a PRs of only $\approx 20$ and 2NN of $\approx 45$.  Because CLIP PMs fill almost all of ambient space, it contains exponentially many images that are confidently assigned by CLIP to any concept, but would not be by humans, as corroborated by visual samples in \cref{fig: clip samples}.

\section{Robustness correlates with alignment}
\label{sec:correlation}

\begin{figure*}[b]
    \centering
    \includegraphics[width=.9\linewidth]{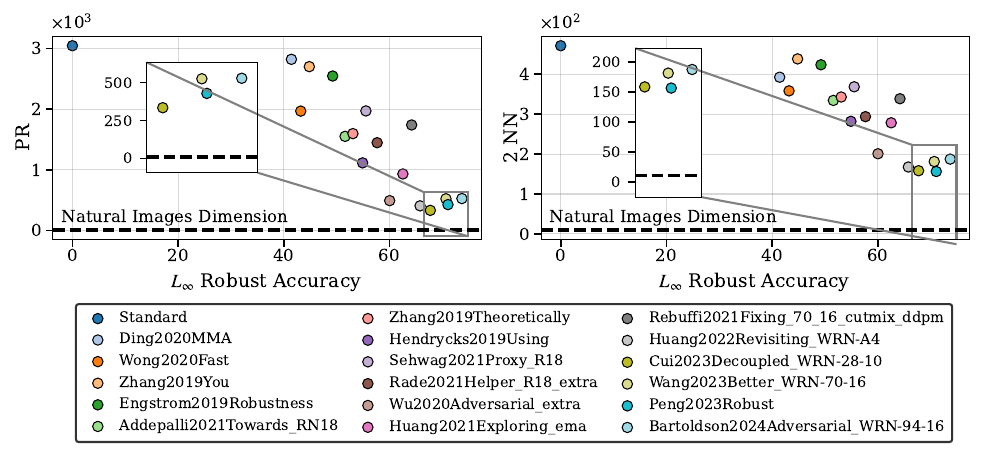}
    \caption{Dimensionality of the Perceptual Manifold (averaged over all CIFAR10 classes) versus $L_\infty$ robust accuracy for models from the CIFAR10 $L_\infty$ model zoo of RobustBench \citep{croce_robustbench_2021}; the models in legend order come from: \citep{ding_mma_2020, wong_fast_2020, zhang_you_2019, madrylab2019robustness, addepalli_scaling_2022, zhang_theoretically_2019, hendrycks_using_2019, sehwag_robust_2022, rade_helper-based_2021, wu_adversarial_2020, huang_exploring_2022, rebuffi_fixing_2021, huang_revisiting_2022, cui_decoupled_2024, wang_better_2023, peng_robust_2023, bartoldson_adversarial_2024}. The left panel displays effective dimensionality (via Participation Ratio), while the right panel displays intrinsic dimensionality (via Two Nearest Neighbors). Note that 2NN values are lower bounds; however, based on the scaling of the predicted intrinsic dimension with dataset size in \cref{fig:scaling 2NN}, the relative ranking will likely stay the same with more samples. We also report the scaling for the PR with samples in \cref{fig:scaling PR}, and the full class by class breakdown in \cref{fig: PR vs LinfAcc all} and \cref{fig: 2NN vs Linfacc all}.}
    \label{fig: dimension vs rob acc avg}
\end{figure*}

Again, our hypothesis is that the existence of successful bounded norm adversarial attacks fundamentally originates from exponential misalignment through the excessive dimensionality of a network's perceptual manifold.  This hypothesis naturally makes two testable predictions. First, more adversarially robust networks should be associated with lower dimensional perceptual manifolds. We successfully test this prediction in Sec.~\ref{subsec:robdim}.  Second, the distance of a random image to the perceptual manifold should increase as the dimensionality of the perceptual manifold decreases, thereby again accounting for the origins of adversarial robustness from reduced dimensionality of perceptual manifolds.  We successfully test this prediction in Sec.~\ref{subsec:dimdist}

\subsection{More robust models have lower dimensional perceptual manifolds}
\label{subsec:robdim}

To test our first prediction, we analyze a set of CIFAR10 classification models from RobustBench \citep{croce_robustbench_2021} that cover a wide range of $L_\infty$ robust accuracies and compute the intrinsic dimensionality of their PMs. We plot the dimensionality as a function of robust accuracy (as reported in the RobustBench leaderboard) in Fig.\ref{fig: dimension vs rob acc avg}. As predicted by our hypothesis, we observe a clear negative correlation between the dimensionality of the PM and robust accuracy, even though none of the models had been explicitly trained to reduce PM dimensionality. We confirm that this inverse correlation between robust accuracy and PM dimensionality scales to high-resolution regimes, observing a similar inverse correlation across robust ImageNet-1K models (see \cref{sec: imagenet}).

Notably, even the most robust models exhibit a PR of $\approx 250$ and 2NN of $\approx 150$, values significantly larger than those of natural images ($\approx 10$ and $\approx 24$, respectively). We explore this persistent misalignment in more detail in Sec.~\ref{sec: persistent misalignment}.
Finally, we observe significant intra-model variation in PM dimensionality across different classes (see App. Fig.~\ref{fig: hetero pr} and App. Fig.~\ref{fig: hetero 2nn}).

% \begin{figure*}[b]
%     \centering
%     \includegraphics[width=.9\linewidth]{images/dim_vs_LinfAcc_average_only_scientific_with_insets_nogrid.pdf}
%     \caption{Dimensionality of the Perceptual Manifold (averaged over all CIFAR10 classes) versus $L_\infty$ robust accuracy for models from RobustBench \citep{croce_robustbench_2021}. The left panel displays effective dimensionality (via Participation Ratio), while the right panel displays intrinsic dimensionality (via Two Nearest Neighbors). Note that 2NN values are lower bounds; however, based on the scaling of the predicted intrinsic dimension with dataset size in \cref{fig:scaling 2NN}, we are confident the relative ranking will stay the same. We also report the scaling for the PR in \cref{fig:scaling PR} and the full class by class breakdown in \cref{fig: PR vs LinfAcc all} and \cref{fig: 2NN vs Linfacc all} }
%     \label{fig: dimension vs rob acc avg}
% \end{figure*}

\subsection{The lower the dimensionality of a perceptual manifold, the further any point is from it}
\label{subsec:dimdist}

\begin{figure*}[t]
    \centering
    \includegraphics[width=0.9\linewidth]{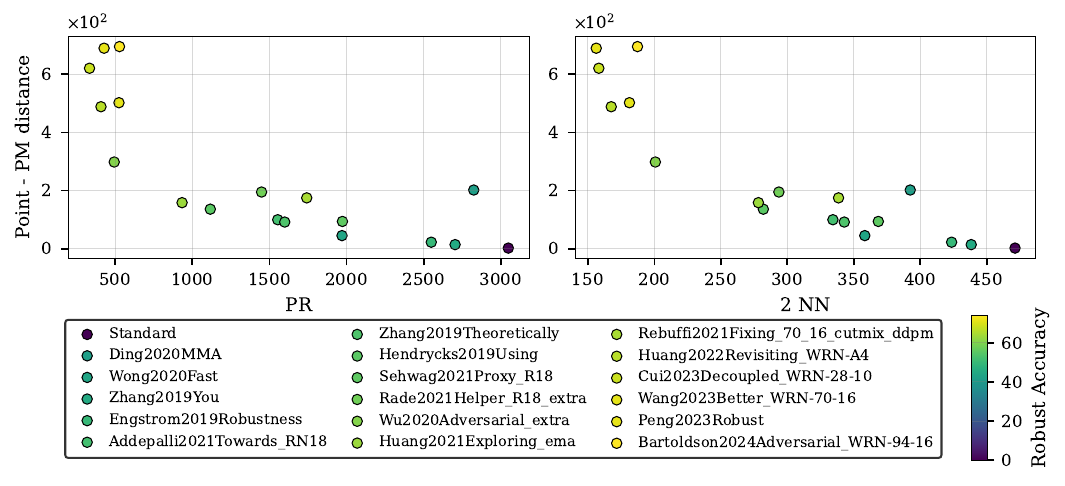}
    \caption{Average squared Euclidean distance between a random point $\mathbf{x} \sim \mathcal{U}[0,1]^D$ and the perceptual manifold, averaged over CIFAR-10 classes. The color scheme indicates robust accuracy (via RobustBench). We observe a strong correlations between robustness, PM dimension and distance to the PM: models with higher robust accuracy exhibit lower PM dimension, and a larger distance to the PM from random noise. For a class-wise breakdown, see \cref{fig:distance vs pr} and \cref{fig:distance vs 2NN}.}
    \label{fig: distance vs dim avg}
\end{figure*}

To test our second prediction, we analyze a wide range of models from RobustBench \citep{croce_robustbench_2021} by measuring the squared Euclidean distance ($L_2^2$) between a random starting image $\mathbf{x} \sim \mathcal{U}[0,1]^D$ sampled uniformly from the ambient space, and the corresponding point reached on the perceptual manifold (PM) via gradient-based optimization. (see Sec.~\ref{subsec:samples} for some resulting samples and App.~\ref{sec: PMsam appendix} for the sampling details). This distance found via opimization yields an upper bound on the shortest possible distance between the initial point and the PM. Fig.~\ref{fig: distance vs dim avg} illustrates the results: we observe a clear trend where this squared distance is largest ($\approx 600$) for the lowest-dimensional PMs and decays towards zero as PM dimensionality increases. This  corroborates the intuition captured by our toy model in \cref{sec:toy_model_distance}: as PM dimension increases, the PM occupies more of the ambient space, thereby significantly reducing the distance required to reach it from any random point. This distance versus dimension scaling relationship holds even when the ambient space expands to over 150,000 dimensions, as demonstrated in our ImageNet experiments (\cref{sec: imagenet}).  We further confirm in App.~\ref{sec: app distance random and image} that the distance to the PM from random noise serves as a reliable proxy for the distance from natural images, the metric most relevant to adversarial robustness.  Together, these results further validate the geometric origins of adversarial vulnerability through exponential misalignment.

% \begin{figure}
%     \centering
%     \includegraphics[width=1\linewidth]{images/from_real_from_random.pdf}
%     \caption{Comparison of distances to the Perceptual Manifold (PM) initialized from natural images versus random noise. The $x$-axis displays the typical squared Euclidean distance from a random point $\mathbf{x} \sim \mathcal{U}[0,1]^D$ to the `truck' class PM; the $y$-axis displays the distance from natural images (excluding the target class) to the same PM. Each point represents a different model. The dashed line denotes the identity $y=x$. For the legend, refer to \cref{fig: distance vs dim avg}}
%     \label{fig: from real vs from random}
% \end{figure}

\section{Adversarial examples are still not solved because exponential misalignment persists}
\label{sec: persistent misalignment}

The insets of Fig.~\ref{fig: dimension vs rob acc avg} reveal that even the most robust networks are still exponentially misaligned on average across all $10$ CIFAR10 classes, with PM dimensionalities far higher than that of natural images.  Here we go beyond dimensionality to explore the nature of this human-machine misalignment at higher resolution, at the level of eigenstructure of PM covariance matrices (App.~\ref{app:alignment_analysis}), and at the level of individual samples from PMs of single networks and classes (Sec.~\ref{subsec:samples}). 

App.~\ref{app:alignment_analysis} reveals that the covariance matrix eigenstructure and power spectral density (PSD) of machine PMs are misaligned to that of natural image classes.  Indeed eigenvectors of PM covariance matrices are not much more aligned to those of natural images, than by random chance. And while more robust models have PSDs closer to that of natural images, they fail to capture their scale invariant $1/f^2$ statistics. 

%To quantify this, we conducted a detailed analysis of the geometry and spectral properties of the Perceptual Manifolds, including their covariance spectra, eigenvector alignment with natural data, and Power Spectral Density (PSD). We present the full detailed analysis in Appendix~\ref{app:alignment_analysis}.
%In summary, our analysis reveals that while robust models exhibit statistical properties closer to natural images than standard models (e.g., steeper PSD slopes and broader covariance spectra), they remain distinct. Crucially, the principal subspaces of the model PMs remain largely orthogonal to the principal directions of natural images, and the PMs fail to fully recover the scale-invariant $1/f^2$ statistics of natural data.
%However, intriguingly, we find that distinct "sparks of alignment" begin to emerge visually in the lowest dimensional PMs.

\subsection{Sparks of alignment in perceptual manifold samples}
\label{subsec:samples}
\begin{figure*}[b!]
    \centering
    \includegraphics[width=0.9\linewidth]{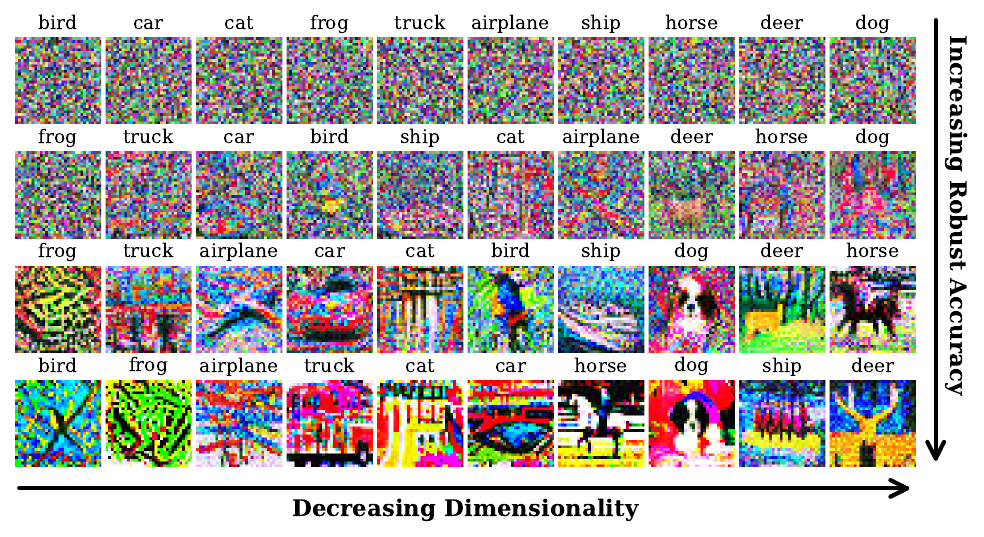}
    \caption{Individual samples from the PMs of all $10$ CIFAR10 classes for each of $4$ different models (1 model per row) with increasing robust accuracy from top to bottom.  In particular, using the same model names as in other figure legends, we show examples from the PMs of  ``Standard", ``Wong2020Fast", ``Wu2020Adversarial$\_$extra", ``Peng2023Robust" with $L_\infty$ robust accuracies from top to bottom of $0\%, 43.21\%, 60.04\%, 71.07\%$ respectively. In each row we have ordered the classes shown according to the PR dimensionality of their PMs for the corresponding network, from highest dimensionality at the left to lowest at the right. We note that we only see alignment between samples of the PM and human perception in the lowest dimensional PMs of the most robust networks.  For example, the recognizable horse, dog and deer samples in the bottom row come from network PMs of PR dimensionality $140$, $133$ and $80$, and 2NN dimensionality $\geq89$, $\geq138$, and $\geq82$ respectively, while the recognizable dog and horse samples in the third row come from network PMs of PR dimensionality $307$ and $289$ and 2NN dimensionality $\geq176$ and $\geq183$ respectively.  In contrast in the second row from the top, despite the network achieving a robust accuracy of $43.21\%$, its PMs are unaligned to human perception with unrecognizable samples because their PM PR and 2NN dimensionalities are all high (ranging from PR$=3016$, and 2NN$\geq470$ for frog to PR$=1397$ and 2NN$\geq295$ for dog). See \cref{fig: PM samples pt 1} and \cref{fig: PM samples pt2} for more samples of PMs from all $18$ networks studied in this paper and all $10$ CIFAR10 classes, again indicating alignment at the level of samples only when PM dimensionality is low.} 
    \label{fig: sparks of alignment}
\end{figure*}

Fig.~\ref{fig: sparks of alignment} shows individual samples from PMs of individual networks of different robustness, and individual class PMs of different dimensionality. See also Fig.~\ref{fig: PM samples pt 1} and~\ref{fig: PM samples pt2} for more samples of PMs from all $18$ networks studied in this paper and all $10$ CIFAR10 classes. A main conclusion from these figures is that whenever the PM dimensionality is much higher than that of natural images (i.e. in almost all cases), random samples from the PM look like noise, indicating exponential misalignment. But intriguingly, for the most robust networks, and their classes with lowest PM dimensionality, approaching that of natural images, we do see a few cases of random samples that look like human recognizable exemplars of the corresponding class.  This indicates that dimensional alignment between human and machine perception can induce sparks of perceptual alignment at the level of individual random samples from the PM. This semantic emergence replicates at scale; in \cref{sec: imagenet}, we show that samples from lower-dimensional ImageNet PMs begin to resolve into distinct, human-recognizable textures and object sub-parts, distinct from the white noise samples of higher dimensional PMs in non-robust models.

\vspace{-1em}
\section{Discussion}

Overall our work defines and centers the notion of perceptual manifolds (PMs) and their dimensionality as central objects in understanding machine perception. This enables us to reveal extreme exponential misalignment between humans and machines, as well as connect this misalignment to the origin of adversarial examples. It would be interesting to extend our work beyond vision to language. There, for any given next token, it is likely that there are exponentially many high probability gibberish token sequences that end in next token. Indeed such sequences can be used to attack LLMs \cite{Yamamura2024-wf}. But more generally, by revealing what we believe to be the fundamental {\it raison d'etre} of adversarial examples, namely exponential misalignment in machine versus human PMs, we hope our work will inspire new ideas for solving adversarial examples once and for all, ideas which may then transfer to solving more general problems of alignment between AI and humans.  
\section*{Impact Statement}

A key issue for the successful societal adoption of AI is alignment of AI to human values and intentions.  By connecting alignment to the notion of adversarial examples, we provide potentially new ways to think about alignment and adversarial examples alike, yielding a cross-fertilization of ideas between these disciplines, and highlighting a common curse of dimensionality as challenge to be overcome in both fields (e.g. achieving good behavior across exponentially many inputs, whether that behavior is perception in the case of adversarial examples, or alignment of values and intentions more generally).

\section*{Acknowledgments}

S.G and A.S thank the Simons Foundation Collaboration on the Physics of Learning and Neural Computation for support. S.G. additionally thanks a Schmidt Sciences Polymath Award for support.

\bibliography{references,references-surya}
\bibliographystyle{icml2026}

%%%%%%%%%%%%%%%%%%%%%%%%%%%%%%%%%%%%%%%%%%%%%%%%%%%%%%%%%%%%%%%%%%%%%%%%%%%%%%%
%%%%%%%%%%%%%%%%%%%%%%%%%%%%%%%%%%%%%%%%%%%%%%%%%%%%%%%%%%%%%%%%%%%%%%%%%%%%%%%
% APPENDIX
%%%%%%%%%%%%%%%%%%%%%%%%%%%%%%%%%%%%%%%%%%%%%%%%%%%%%%%%%%%%%%%%%%%%%%%%%%%%%%%
%%%%%%%%%%%%%%%%%%%%%%%%%%%%%%%%%%%%%%%%%%%%%%%%%%%%%%%%%%%%%%%%%%%%%%%%%%%%%%%
\newpage
\appendix
\onecolumn

\iffalse

\section{Potential weaknesses / doubts  (obviously won't be in final paper)}

\subsection{Phenomena we don't explain}
\begin{itemize}
    \item Transferrability 
    \item Robustness Accuract Trade Off
\end{itemize}

\section{Other Analysis that might be worth doing (obviously won't be in final paper)}
\begin{itemize}
    \item Should I do the same ``spectral analysis" as done in \cref{sec: persistent misalignment}, but for CLIP? It is computationally very expensive. If not, maybe it is better to put the CLIP section before the "Robustness correlates with alignment", as we do the same things as in "Fragility from exponential misalignment", but for CLIP
    \item Classes with lower dimensional PM should be harder to have as the target one in targeted adversarial attacks
\end{itemize}

\fi

\section{Background and Related Work} \label{sec: Background and Related Work}
The existence of adversarial attacks, first noted by \citep{szegedy_intriguing_2014}, remains a highly intriguing property of neural networks: highly performant models on a test set can still nevertheless be fooled by small input perturbations that are imperceptible to humans.  Many possible attacks and defenses have been proposed in the literature, and many theories have been proposed for why they exist. Still, no consensus has yet been reached on why they exist, nor has a strong universal defense strategy been found. Since our primary focus is a new framework to explain why adversarial examples exist (namely high dimensionality of machine perceptual manifolds) and what might be required to remove adversarial examples (namely lowering this dimensionality), in our discussion of related works we focus on various proposed theories for adversarial examples and how they may relate to our proposal. 

\textbf{Linearity and Geometry.} The ``linear theory'' proposed by \citep{goodfellow_explaining_2015} was the first to posit that adversarial examples arise from the excessive linearity of neural networks in high-dimensional spaces. While foundational, this hypothesis struggles to explain the superior efficacy of iterative attacks (e.g., PGD \citep{madry_towards_2019}) over single-step methods (e.g., FGSM), suggesting that local non-linearity is a critical factor. \citep{tanay_boundary_2016} provide several arguments against the linearity hypothesis as a sufficient explanation and argue instead  for a geometric explanation where decision boundaries ``tilt'' relative to the data manifold, lying close to data points along directions of low variance.

\textbf{Feature Alignment and Non-Robust Features.} \citep{ilyas_adversarial_2019} reframe adversarial examples as a different form misalignment phenomenon, suggesting models rely on ``non-robust features''—signals that are highly predictive but imperceptible to humans. While empirically robust, we argue that accepting reliance on these features is a concession that retreats from attempting to align the algorithms of machine vision with those of human vision. Furthermore, if adversarial vulnerability were driven solely by the existence of these features, it would not explain the geometric collapse we observe: we find that robust models do not merely ignore specific features, but fundamentally compress the dimensionality of their {\it entire} perceptual manifold (PM).

\textbf{High Dimensions and Concentration of Measure.} A significant body of work posits that adversarial examples are an inevitable consequence of high dimensionality of the input data manifold \citep{gilmer_adversarial_2018}, ambient space \citep{shafahi_are_2020, fawzi_adversarial_2018} and non-zero test error \citep{ford_adversarial_2019, mahloujifar_curse_2018}. While we agree that high dimensionality is the culprit, we propose a distinct, more actionable thesis. Previous works often focused on the \textit{ambient} dimension or the \textit{data} dimension. We demonstrate that the critical factor is neither, but rather the dimensionality of the \textit{machine's perceptual manifold}. Unlike the ambient dimension (which is fixed) or data dimension (which is low), the PM dimension is the property that correlates directly with robustness and which, more importantly, can be acted upon and reduced via alternate training methods. 

\textbf{Manifold Hypothesis.} Another branch of papers focuses on the manifold hypothesis as the cause of adversarial examples. These works propose that adversarial examples exist because the model behaves robustly on manifold, but as soon as one steps outside of it the model becomes extremely brittle since it hasn't seen any samples in that region. Notable works in this direction are \citep{khoury_geometry_2018}, which focuses on the discrepancy between manifold and ambient dimension, and \citep{ma_characterizing_2018}, which uses the Local Intrinsic Dimension (LID) to distinguish between regular (on manifold) and adversarial (off manifold) samples. This however can't be the full explanation as \citep{stutz_disentangling_2019} finds that adversarial examples exist also on manifold. While our theory also has out of manifold adversarial examples as a natural consequence, it is not limited to that. 

\textbf{Sample Complexity.} \citep{schmidt_adversarially_2018} establish an information-theoretic lower bound demonstrating that adversarially robust generalization requires significantly larger sample complexity than standard generalization, specifically scaling as $\Omega(\sqrt{d})$ in high-dimensional Gaussian models. This reflects in practice in the success of augmenting the training set with diffusion models as a defense strategy. This can be seen as a way to align the models' PM with the actual data distribution, hence in line with our work. 

\textbf{Topology of Decision Boundaries.} The topology of the decision boundary and its effect on adversarial robustness has been studied extensively, though only in the local proximity of the dataset and never in its global dimensionality. \citep{fawzi_empirical_2018} characterize decision regions as connected but non-convex, showing that the curvature of the boundary can be used to distinguish adversarial examples. Similarly, \citep{nguyen_neural_2018} posit that the insufficient expressivity of standard models forces them to learn connected decision regions, a topological constraint that inevitably gives rise to adversarial examples.

\section{Sampling the Perceptual Manifold}
\label{sec: PMsam appendix}
Here we report the basic algorithm to sample the PM. We note that, for the optimization to be successful, it is sometimes necessary to either switch to Adam (especially for CLIP \citep{radford_learning_2021}), or to add noise and/or momentum (this mostly happens for more robust models). We also note that we weren't able to successfully complete the optimization for two model-class pairs, namely class 5 for \citep{cui_decoupled_2024} (``Cui2023Decoupled\_WRN02810") and class 7 for \cite{bartoldson_adversarial_2024}(``Bartoldson2024Adversarial\_WRN-94-16"). Moreover we note that all those models that use some form of gradient masking were avoided as they would have made sampling much more computationally heavy.

\begin{algorithm}[h]
\caption{Projected Gradient Ascent}
\label{alg:pga_sampling}
\begin{algorithmic}[1]
% Inputs
\REQUIRE Classifier $f(x)$ (outputting logits), Target class $c$, Step size $\alpha$, Probability threshold $\tau$, Max iterations $T_{\max}$
% Outputs
\ENSURE Optimized input $x^*$

\STATE $x_0 \leftarrow \text{Sample } \mathcal{U}[0, 1]^D$
\STATE $t \leftarrow 0$

% We need to compute the probability for the threshold check
\STATE $p_0 \leftarrow \text{Softmax}(f(x_0))$

\WHILE{$p_{t,c} < \tau$ \textbf{and} $t < T_{\max}$}
    % We maximize log(probability)
    \STATE $g_t \leftarrow \nabla_x \log (\text{Softmax}(f(x_t))_c)$ \COMMENT{Ascent direction}
    \STATE $\hat{x}_{t+1} \leftarrow x_t + \alpha \cdot g_t$
    \STATE $x_{t+1} \leftarrow \text{clip}(\hat{x}_{t+1}, 0, 1)$
    \STATE $p_{t+1} \leftarrow \text{Softmax}(f(x_{t+1}))$ \COMMENT{Update prob}
    \STATE $t \leftarrow t + 1$
\ENDWHILE

\STATE \textbf{return} $x_t$ 
\end{algorithmic}
\end{algorithm}

\section{Estimating the intrinsic dimension of a perceptual manifold from samples} \label{sec:estindim}

The 2 nearest neighbor (2NN) method \citep{facco_estimating_2017} estimates the \textit{intrinsic dimension} of a manifold, i.e. the minimum number of degrees of freedom required to traverse the manifold. It is derived from the statistics of local neighbor distances: for any sample, define $\mu = r_2/r_1$ as the ratio of distances to the second ($r_2$) and first ($r_1$) nearest neighbor samples. If points are sampled from a manifold of intrinsic dimension $d$, then this ratio follows a Pareto distribution with cumulative distribution function (CDF) $F(\mu) = 1 - \mu^{-d}$, which linearizes in logarithmic coordinates as $\log(1 - F(\mu)) = -d \log(\mu)$. To estimate $d$ from samples we then compute the ratios $\{\mu_i\}_{i=1}^N$ for all $N$ samples, sort them in ascending order, and approximate the CDF via the empirical rank $F_{\text{emp}}(\mu_{(i)}) \approx i/N$. The dimension $d$ is then estimated from a linear regression through the origin over $i=1, \dots, N-1$:
\begin{equation}
d_{\text{2NN}} = -\frac{\sum_{i=1}^{N-1} \log\mu_{(i)} \log(1 - i/N)}{\sum_{i=1}^{N-1} (\log\mu_{(i)})^2}.
\end{equation}
However, the curse of dimensionality requires the sample size $N$ to scale exponentially with $d$ to ensure the manifold is sufficiently densely sampled to accurately estimated $d$. For finite datasets, boundary effects dominate, leading to a systematic negative bias in the estimate of $d$. Consequently, the reported values should be interpreted as lower bounds on the true intrinsic dimension.  Nevertheless we will find that our estimated lower bounds on intrinsic dimension of machine PMs still far exceed the dimension of natural image concepts, for which instead the samples are enough to estimate the intrinsic dimension reliably.

\section{Adversarial attacks and robust accuracy}
\label{sec:advattrobacc}

Generating untargetted adversarial attacks  requires solving a constrained optimization problem over the non-convex loss landscape defined by the network:
\begin{equation}
\max_{\delta \in \mathcal{S}} \mathcal{L}(g(x+\delta), c).
\end{equation}
A canonical method for solving this is PGD \citep{madry_towards_2019}, from which we take inspiration for our sampling algorithm \cref{subsec:PMsam}. PGD iteratively maximizes a surrogate loss function, typically the Cross-Entropy loss or the Carlini-Wagner loss \citep{carlini_towards_2017} defined on logit differences, by taking steps in the direction of the gradient and projecting back onto the feasible set $\mathcal{S}$. The update rule at iteration $k$ is given by:
\begin{equation}
x^{(k+1)} = \Pi_{\mathcal{S}}\left(x^{(k)} + \eta \cdot \text{sign}(\nabla_x \mathcal{L}_{\text{CE}}(g(x^{(k)}), c))\right),
\end{equation}
where $\eta$ is the step size and $\Pi_{\mathcal{S}}$ denotes the projection onto the $\epsilon$-ball.
To quantify vulnerability, we utilize the RobustBench leaderboard \citep{croce_robustbench_2021}, which employs the AutoAttack ensemble \citep{croce_reliable_2020} to establish a reliable baseline for worst-case performance. We emphasize that the robust accuracy values reported in this benchmark—and consequently adopted in our work—are based on the untargeted threat model. However, we note that also a targeted version of the attacks exists, although less common, which seeks to force the model to misclassify the input into a specific target class $t$ rather than simply causing a prediction error.

In response to these threats, Adversarial Training (AT) \citep{madry_towards_2019} has established itself as the dominant defense framework. By incorporating adversarial examples into the training set, models learn to be resilient against worst-case perturbations. The current state-of-the-art on the RobustBench leaderboard extends this approach by leveraging synthetic data generated via diffusion models \citep{wang_better_2023} or by utilizing regularization terms like TRADES \citep{zhang_theoretically_2019}, which penalize the divergence between predictions on clean and perturbed inputs to stabilize the decision boundary. Beyond standard AT, other notable strategies include diffusion purification \citep{nie_diffusion_2022}, which aims to project adversarial inputs back onto the data manifold prior to classification, and randomized smoothing \citep{cohen_certified_2019}, which constructs a provably robust classifier by averaging predictions over Gaussian noise.

\section{Derivation of Distance in Toy Model}
\label{sec:toy_model_distance}

We derive the expected squared Euclidean distance between a random query point $x_0 \sim \mathcal{U}[0,1]^D$ and the class manifold $\mathcal{E}$, modeled as a $d$-dimensional filled ellipsoid embedded in $\mathbb{R}^D$.

\paragraph{Setup and Decomposition.}
Let the manifold $\mathcal{E}$ be centered at $c$, with principal semi-axes $r_1, \dots, r_d$. We define the random difference vector $z = x_0 - c$. Assuming $x_0$ and $c$ are independent and uniformly distributed in the unit hypercube, $z$ has zero mean and covariance:
\begin{equation}
\Sigma_z = \text{Cov}(x_0) + \text{Cov}(c) = \frac{1}{6} I_D.
\end{equation}
The squared distance decomposes orthogonally into a perpendicular component $d_\perp^2$ and a parallel component $d_\parallel^2$:
\begin{equation}
\text{dist}(x_0, \mathcal{E})^2 = d_\perp^2 + d_\parallel^2.
\end{equation}

\paragraph{Perpendicular Component.}
The perpendicular distance corresponds to the projection of $z$ onto the subspace orthogonal to the manifold. Let $P_\perp$ be the projection operator onto the $(D-d)$ orthogonal dimensions. Using the cyclic property  of the trace and $\mathbb{E}\left[z\right]=0$, we get $\mathbb{E}[z^\top A z] = \text{Tr}(A \Sigma_z)$. Therefore we obtain the exact result:
\begin{equation}
\mathbb{E}[d_\perp^2] = \text{Tr}\left( P_\perp \frac{1}{6} I_D \right) = \frac{D - d}{6}.
\end{equation}

\paragraph{Parallel Component: Concentration and Effective Radius.}
Let $y$ be the projection of $z$ onto the $d$-dimensional manifold subspace. Using the same reasoning as for the perpendicular component, its expected squared norm is $\mathbb{E}[\|y\|^2] = \text{Tr}(I_d \Sigma_z) = d/6$.
In high dimensions ($d \gg 1$), the \textbf{concentration of measure} phenomenon implies that the random vector $y$ lies in a thin spherical shell. Its radial length $r = \|y\|$ concentrates sharply around the root-mean-square (RMS) value:
\begin{equation}
r \approx \sqrt{\mathbb{E}[\|y\|^2]} = \sqrt{\frac{d}{6}}.
\end{equation}
Fluctuations around this radius are $O(1)$ and become negligible relative to the mean as $d$ increases.

To approximate the parallel distance, we model the ellipsoid as a sphere. We choose its effective radius to be the root mean square of the principal axes $R_{\text{eff}}=\sqrt{\frac{1}{d}\sum r_i^2}$ to preserve the expected square norm of a uniform point drawn from the ellipsoid. 
% We consider two candidates for $R_{\text{eff}}$:
% \begin{enumerate}
%     \item \textbf{RMS Radius:} We match the statistical properties relevant to squared distances. Equating the expected squared norm of a uniform point in the ellipsoid to that of a ball yields $R_{\text{rms}} = \sqrt{\frac{1}{d}\sum r_i^2}$. I think this might be better since we are computing the squared distance.
%     \item \textbf{Geometric Mean Radius:} To preserve the volume of the original ellipsoid, we may choose $R_{\text{geo}} = (\prod r_i)^{1/d}$.
% \end{enumerate}
Using the concentration of measure assumption, the parallel distance is simply the radial excess of the query vector $y$ beyond this effective boundary.
\begin{equation}
\mathbb{E}[d_\parallel^2] \approx \left( \max\left\{0, \sqrt{\frac{d}{6}} - R_{\text{eff}} \right\} \right)^2.
\end{equation}

% \paragraph{Validation via Gaussian Surrogate (Gamma Functions). [LLM Assisted]}
% To validate the concentration heuristic, we performed a more rigorous calculation by approximating the projection $y$ as a Gaussian vector $y \sim \mathcal{N}(0, \sigma^2 I_d)$ with $\sigma^2 = 1/6$. This is a tractable surrogate that preserves the exact second moment $\mathbb{E}[\|y\|^2] = d\sigma^2$.
% Under this assumption, the radial density $f_r(r)$ is known in closed form:
% \begin{equation}
% f_r(r) = \frac{r^{d-1}}{2^{\frac d2-1}\Gamma(\tfrac d2)\sigma^d} \, e^{-r^2/(2\sigma^2)}, \quad r \ge 0.
% \end{equation}
% The expected parallel distance is the integral of the squared radial excess $(r - R)^2$ over the region where $r > R$:
% \begin{equation}
% \mathbb{E}[d_\parallel^2] = \int_R^\infty (r^2 - 2Rr + R^2) f_r(r) \, dr.
% \end{equation}
% By introducing the dimensionless cutoff $s_0 = R^2/(2\sigma^2)$, each term in this expansion reduces to a standard Gamma-type tail integral. The exact solution is:
% \begin{equation}
% \mathbb{E}[d_\parallel^2] = \frac{1}{\Gamma(\tfrac d2)} \Big[ 2\sigma^2\,\Gamma(\tfrac d2+1, s_0) - 2R\sqrt{2}\sigma\,\Gamma(\tfrac{d+1}{2}, s_0) + R^2\,\Gamma(\tfrac d2, s_0) \Big].
% \end{equation}
% Comparison with numerical simulations shows that the simple concentration approximation (Eq. 7) captures the exact same behavior as this rigorous integration in the high-dimensional limit, justifying the use of the simpler expression.

\paragraph{Final Result.}
Combining the exact perpendicular contribution with the concentrated parallel contribution yields:
\begin{equation}
\mathbb{E}[\text{dist}(x_0, \mathcal{E})^2] \approx \frac{D - d}{6} + \left( \max\left\{0, \sqrt{\frac{d}{6}} - \sqrt{\frac{1}{d}\sum_{i=1}^d r_i^2} \right\} \right)^2.
\end{equation}

\section{The distance from random point is a good proxy for the distance from images}
\label{sec: app distance random and image}
\begin{figure}[H]
    \centering
    \includegraphics[width=.5\linewidth]{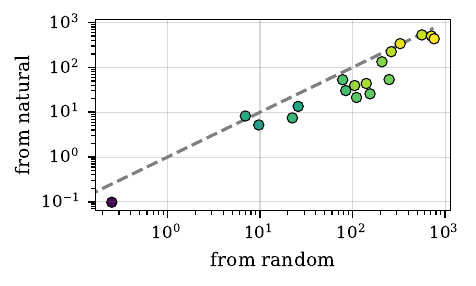}
    \caption{Comparison of distances to the Perceptual Manifold (PM) initialized from natural images versus random noise. The $x$-axis displays the typical squared Euclidean distance from a random point $\mathbf{x} \sim \mathcal{U}[0,1]^D$ to the `truck' class PM; the $y$-axis displays the distance from natural images (excluding the target class) to the same PM. Each point represents a different model. The dashed line denotes the identity $y=x$. For the legend, refer to \cref{fig: distance vs dim avg}}
    \label{fig: from real vs from random}
\end{figure}

To validate whether the distance from random noise serves as a reliable proxy for the distance from natural images—the metric most relevant to adversarial robustness—we repeated the sampling procedure using natural images as initialization points. \Cref{fig: from real vs from random} compares the distances obtained from natural image initialization against those from random noise initialization across various models. We observe a strong correlation between the two metrics, with the distance from a natural image being consistently slightly lower. This indicates that the arguments developed for random noise generalize to natural data.

\section{Geometric and Spectral Analysis of the Perceptual Manifold}
\label{app:alignment_analysis}

In this section, we provide a deeper quantitative analysis of the misalignment between the Perceptual Manifold (PM) and natural data, focusing on covariance spectra, subspace alignment, and frequency statistics.

\subsection{Misalignment in perceptual manifold spectra}
\label{app:spectrum}

\begin{figure}[H]
    \centering
    \includegraphics[width=0.5\linewidth]{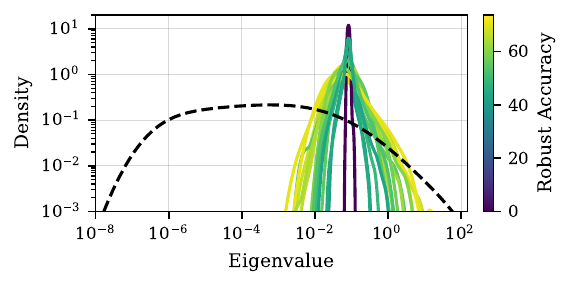}
    \caption{Full colored lines: density of the spectrum of the covariance matrix of the PM of different model. Dashed black line: density of the spectrum of the covariance of natural images. Both refer to class 9 of CIFAR10, i.e. trucks. For the plots of all classes refer to \cref{fig: eigenvalue distribution all classes}}
    \label{fig:covariance spectrum comparison}
\end{figure}

In \cref{fig:covariance spectrum comparison}, we compare the spectral density of the covariance matrices for the model PMs against that of natural images. The plot reveals a stark geometric misalignment. The natural image spectrum (dashed line) exhibits a broad, heavy-tailed distribution extending to extremely small eigenvalues ($< 10^{-6}$), characteristic of a manifold with low intrinsic dimensionality where variance is collapsed in most directions. 

In contrast, standard non-robust models (represented by dark purple lines) exhibit a sharp, narrow peak around $10^{-1}$, indicative of a highly isotropic, high-dimensional geometry where variance is distributed roughly equally across all directions. As robust accuracy increases (transitioning to yellow), the PM spectrum broadens significantly. This broadening indicates that the manifold is becoming progressively more anisotropic, compressing in certain directions to better approximate the data geometry. However, even the most robust models fail to recover the long tail of vanishing eigenvalues inherent to natural images, confirming that significant misalignment persists.

\subsection{Misalignment in perceptual manifold eigenvectors}
\label{app:eigenvec}

To investigate whether the PM of robust models aligns more closely with the natural image space, we define an \textit{Alignment Score}: the average cosine of the principal angles between the subspace spanned by the top $k$ eigenvectors of natural images (sufficient to explain 90\% of variance) and the top $m$ eigenvectors of the model's PM. 

\begin{figure}[H]
    \centering
    \begin{subfigure}[b]{0.45\textwidth} 
        \centering
        \includegraphics[width=\linewidth]{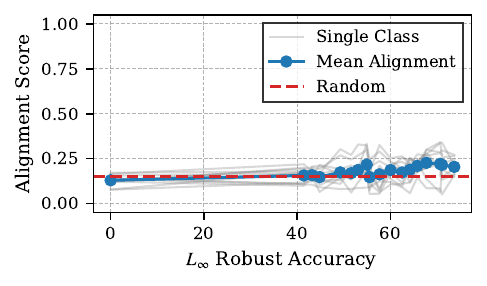}
        \caption{Fixed dimension comparison ($k=m$).}
        \label{fig:alignment fixed k}
    \end{subfigure}
    \hfill
    \begin{subfigure}[b]{0.51\textwidth} 
        \centering
        \includegraphics[width=\linewidth]{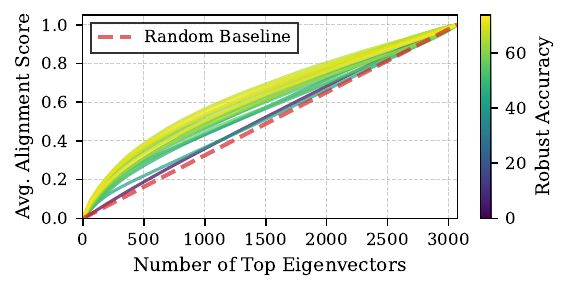}
        \caption{Varying model subspace dimension $m$.}
        \label{fig:alignment increasing k}
    \end{subfigure}
    \caption{Alignment Score analysis between natural image subspaces and Model PM subspaces. $k$ is defined as the number of eigenvectors required to explain $90\%$ of the variance in natural images. (a) Comparison with fixed $k=m$. We observe a very small increase in alignment as robust accuracy improves, though values remain close to the random baseline (red dashed line). (b) Alignment as the model subspace dimension $m$ increases. Even robust models (yellow) fail to capture the natural subspace significantly faster than random chance.}
    \label{fig:alignment_analysis}
\end{figure}

The results, illustrated in \cref{fig:alignment_analysis}, show that misalignment is still heavily present in all models. In \cref{fig:alignment fixed k}, where we fix $m=k$, we observe a subtle trend: alignment correlates positively with robustness, rising from values compatible with chance in standard models to values slightly above the random baseline in the most robust ones. However, despite this relative improvement, the absolute alignment remains low, indicating that the primary directions of variance in the Perceptual Manifold remain largely distinct from the principal components of natural images.

In \cref{fig:alignment increasing k}, we relax the constraint and sweep the model subspace dimension $m$ from $k$ to the full ambient dimension ($D=3072$). While robust models (yellow curves) exhibit a marginal spectral overlap improvement over non-robust ones (purple curves), they fail to concentrate natural signal in their top eigenvectors. Instead, the alignment score grows almost linearly with $m$, tracking the random baseline. This confirms that the robust PM remains rotationally misaligned, with natural features scattered across thousands of dimensions rather than compressed into the principal subspace.

\subsection{Power Spectral Density Analysis of PM}
\label{app:psd}

\begin{figure}[H]
    \centering
    \includegraphics[width=0.5\linewidth]{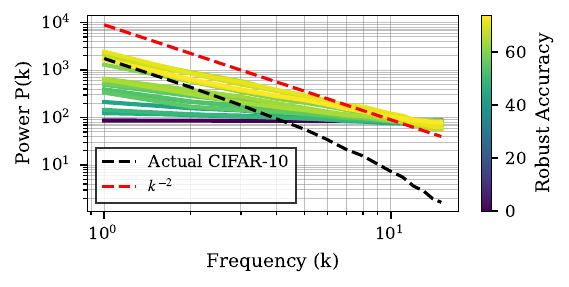}
    \caption{Radially averaged Power Spectral Density (PSD) of PM samples compared to natural CIFAR-10 images. The PSD is computed for each class separately and then averaged. Note the log-log scale: robust models show a steeper slope closer to natural images, but higher overall power. Black dashed line is the PSD of actual CIFAR10 images, red dashed line is the expected $k^{-2}$ line}
    \label{fig:psd avg}
\end{figure}

To quantify the "naturalness" of the samples drawn from the Perceptual Manifold and the gradual alignment between robust models and humans, we analyze their Power Spectral Density (PSD). A fundamental statistical property of natural images is scale invariance, which manifests in the frequency domain as a power law $P(k) \propto 1/k^\alpha$, where $k$ is the spatial frequency and typically $\alpha \approx 2$ \citep{ruderman_statistics_nodate}. In \cref{fig:psd avg}, we plot the radially averaged PSD of samples generated from models with varying degrees of robustness and compare them to natural CIFAR-10 images.

\begin{itemize}
    \item \textbf{Standard Models:} The spectrum is effectively flat ($\alpha \approx 0$), indicating that the PM for non-robust models is effectively white noise, lacking spatial correlations.
    \item \textbf{Robust Models:} As robust accuracy increases, we observe a progressive steepening of the spectral slope. The samples distinctively deviate from the flat spectrum of standard models, trending towards the characteristic $1/k^2$ decay of natural statistics, although they do not fully recover the natural profile.
\end{itemize}

Additionally, we observe that the overall power (magnitude) of the PM samples is consistently higher than that of natural images. We attribute this energy excess to two factors: (1) the geometry of the high-dimensional hypercube, where the volume is concentrated near the corners (vertices), naturally biasing samples toward extreme pixel values; and (2) the gradient-based maximization used for sampling, which tends to saturate features to maximize class probability.

This analysis confirms that despite the partial compression of the manifold towards natural signal statistics observed in robust models, the alignment remains incomplete.

\section{ImageNet Experiments}
\label{sec: imagenet}

To verify that the results reported in the main text are not an artifact of low-resolution datasets like CIFAR-10, we replicated our core analyses on ImageNet-1K \citep{russakovsky_imagenet_2015}. We once again utilized pretrained models from the RobustBench leaderboard \citep{croce_robustbench_2021}, restricting our analysis to models with a clean accuracy of $\geq 70\%$. This constraint isolates the geometric effects of robustness from general baseline (clean) performance differences. Due to the severe computational constraints of high-dimensional optimization, we exclusively use the Participation Ratio (PR) to estimate dimensionality and set our target confidence threshold to $p_0 = 0.8$, once again higher than that of robust networks ($60\%-70\%$ on correctly classified images) and comparable to that of standard non robust networks (often $\geq 85\%$).

\subsection{Exponential Misalignment at Scale}
\label{subsec: imagenet exponential misalignment}

\begin{figure}[h]
    \centering
    \includegraphics[width=0.5\linewidth]{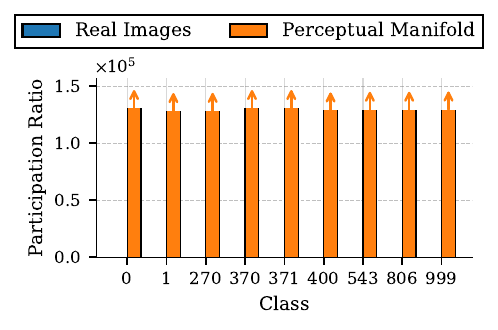}
    \caption{Comparison of the Participation Ratio (PR) of the Perceptual Manifold for a standard ResNet-50 (clean accuracy of 76.52\%, robust accuracy of 0\%) against natural images across $9$ different ImageNet classes. Note that the estimated PR of the perceptual manifold represents a lower bound. The PR of natural images is $\approx 20$, rendering it virtually imperceptible compared to the $\geq 130,000$ dimensions occupied by the machine PM.}
    \label{fig: imagenet exponential misalignment}
\end{figure}

Exponential misalignment—the severe dimensional mismatch between human and machine perceptual manifolds—persists and is arguably more pronounced in standard CNNs trained on ImageNet. As reported in \cref{fig: imagenet exponential misalignment}, the natural image manifold for a given ImageNet class exhibits a PR of approximately 20, consistent with the intrinsic dimensionality observed in CIFAR-10. In stark contrast, the PM of a standard ResNet-50 (76.52\% clean accuracy, 0\% robust accuracy) expands to fill nearly the entire ambient space, occupying over 130,000 of the 150,528 available dimensions. This represents a dimensional mismatch of nearly four orders of magnitude, confirming that standard training paradigms consistently learn class representations that densely permeate the input space.

\subsection{Robustness Correlates with Dimensional Compression}

\begin{figure}[h]
    \centering
    \includegraphics[width=0.7\linewidth]{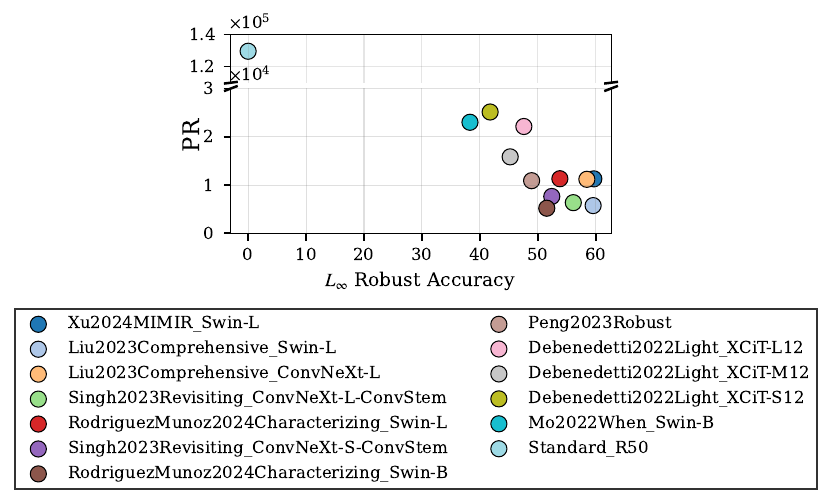}
    \caption{Participation Ratio of the Perceptual Manifold (averaged over 9 ImageNet classes) versus $L_\infty$ robust accuracy for models from the RobustBench ImageNet model zoo \citep{croce_robustbench_2021}. The evaluated models, in legend order, are from \citep{xu_mimir_2025, liu_comprehensive_2023, singh_revisiting_2023, rodriguez-munoz_characterizing_2024, peng_robust_2023, debenedetti_light_2023, mo_when_2022}. Full class-by-class breakdowns and dataset size scaling are provided in \cref{fig:PR_vs_robacc_by_class_imagenet} and \cref{fig:PR_robac_scaling_imagenet}, respectively.}
    \label{fig:PR-vs_RobAcc_imagenet}
\end{figure}

\begin{figure}[h]
    \centering
    \includegraphics[width=0.7\linewidth]{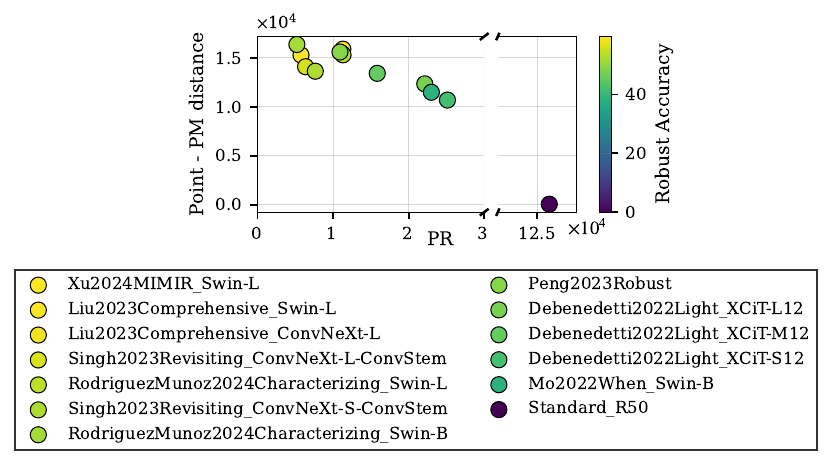}
    \caption{Average squared Euclidean distance between a random initialization $x\sim\mathcal{U}[0,1]^D$ and the Perceptual Manifold as a function of the Participation Ratio, averaged over 9 ImageNet classes. Node colors indicate $L_\infty$ robust accuracy as reported by RobustBench. See \cref{fig:distance_vs_pr_all_classes_imagenet} for class-wise breakdowns.}
    \label{fig:distance_vs_pr_imagenet}
\end{figure}

We next evaluated whether the inverse correlation between adversarial robustness and PM dimensionality holds at the ImageNet scale. \Cref{fig:PR-vs_RobAcc_imagenet} demonstrates a clear downward trend in the PM Participation Ratio as $L_\infty$ robust accuracy increases across the selected model zoo. Furthermore, as the PM dimensionality collapses, the expected squared Euclidean distance from a uniform random noise initialization to the PM strictly increases (\cref{fig:distance_vs_pr_imagenet}). This confirms our geometric hypothesis at scale: reducing the dimensionality of the PM shrinks its footprint in the ambient space, pushing it further away from arbitrary points and naturally increasing the minimum perturbation norm required for an adversarial attack. 

\subsection{Sparks of Alignment in High-Resolution Samples}

\begin{figure}
    \centering
    \includegraphics[width=\linewidth]{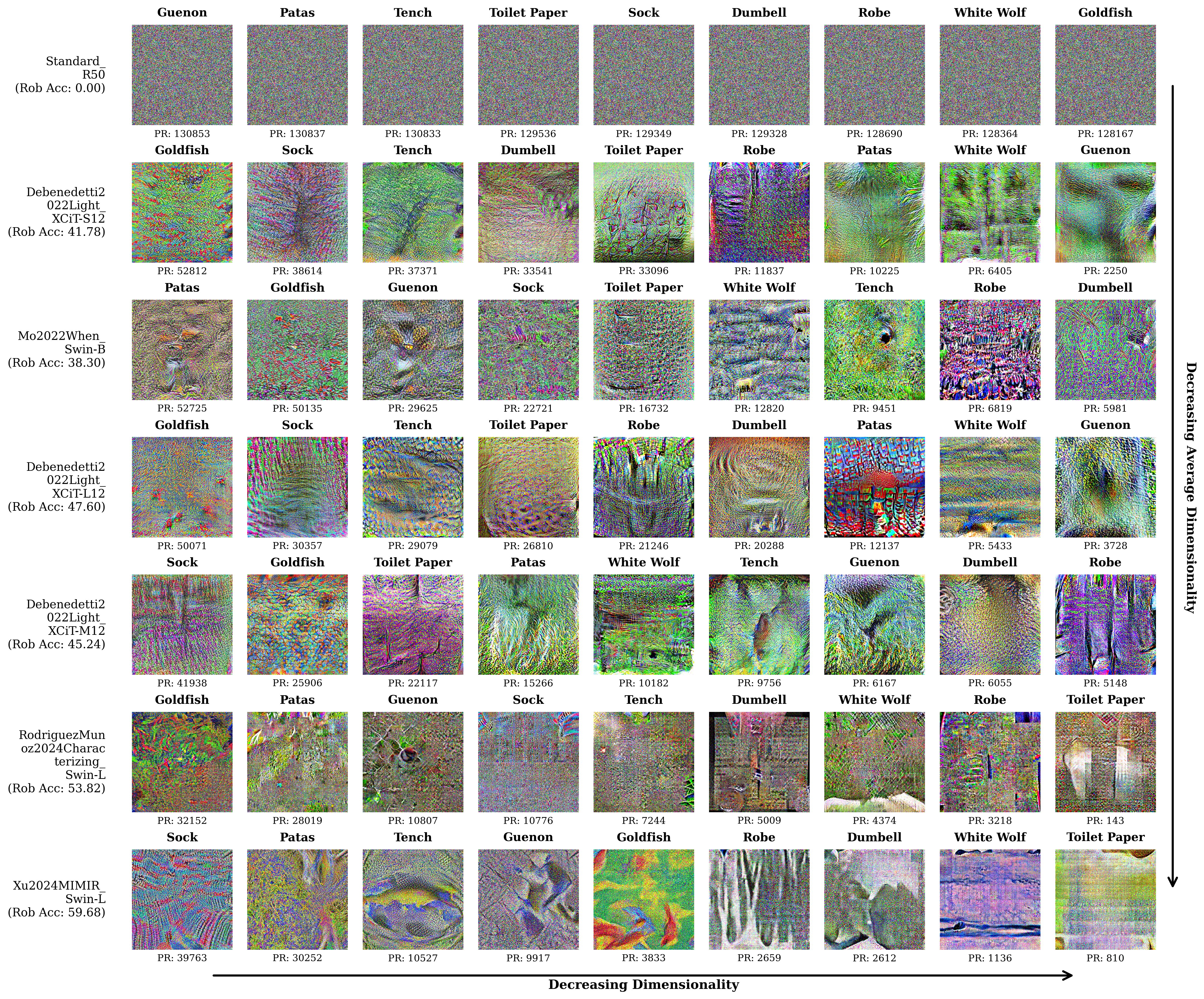}
    \caption{pt 1. Samples from the PM of 9 ImageNet classes across the 7 models whose PM have the highest average (over classes) dimensionality. Rows are sorted in descending order top to bottom by the model's average PR, which is correlated with the robust accuracy. Columns (classes) are sorted by the PR of the corresponding PM descending from left to right.}
    \label{fig:all_images_sorted_by_avg_pr_part1_imagenet}
\end{figure}

\begin{figure}
    \centering
    \includegraphics[width=\linewidth]{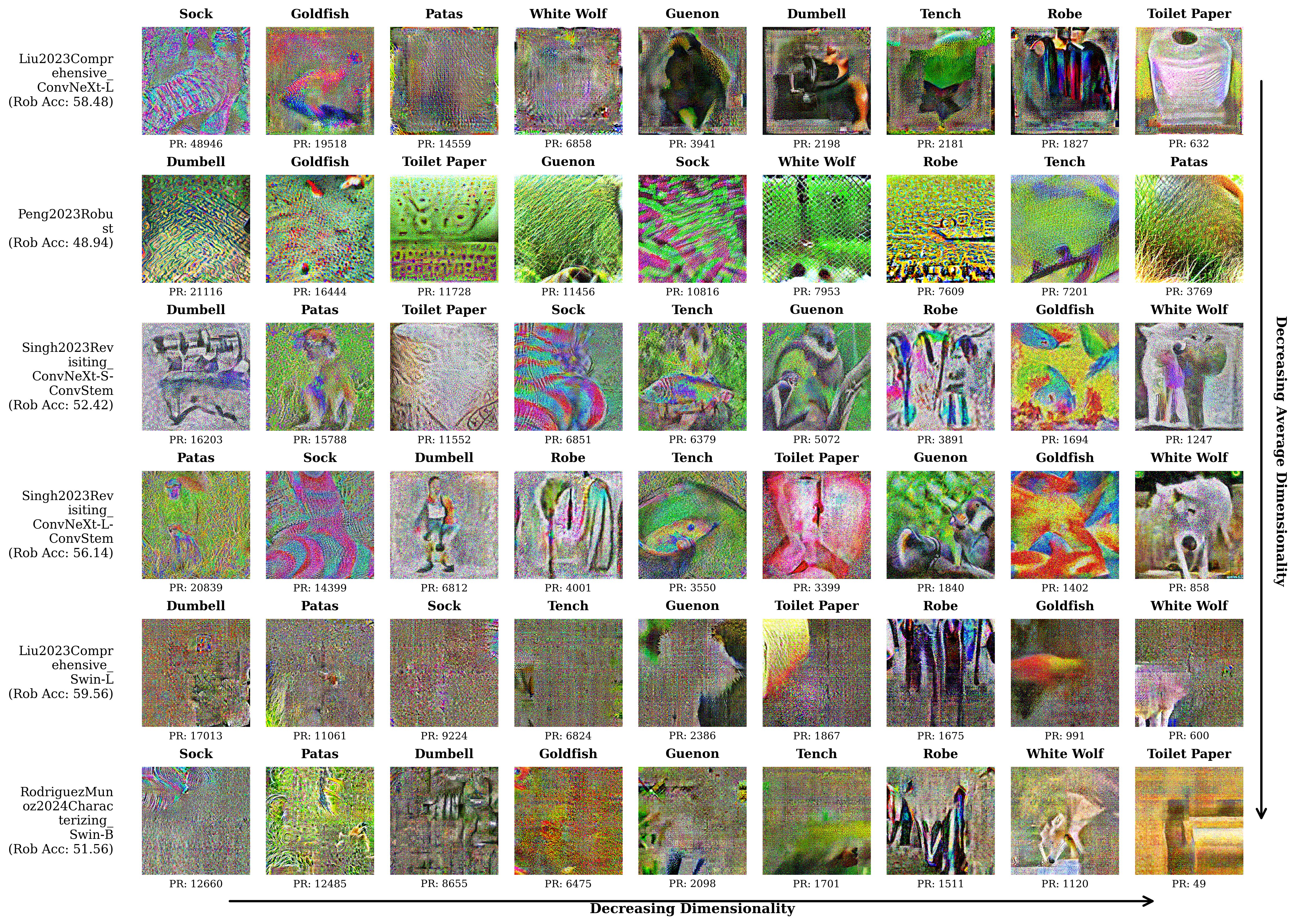}
    \caption{pt 2. Samples from the PM of 9 Imagenet classes across the 6 models whose PM have the lowest average (over classes) dimensionality. Rows are sorted in descending order top to bottom by the model's average PR, which is correlated with the robust accuracy. Columns (classes) are sorted by the PR of the corresponding PM descending from left to right.}
    \label{fig:all_images_sorted_by_avg_pr_part2_imagenet}
\end{figure}

Visual inspection of the ImageNet PM samples (\cref{fig:all_images_sorted_by_avg_pr_part1_imagenet} and \cref{fig:all_images_sorted_by_avg_pr_part2_imagenet}) corroborates the semantic convergence observed in \cref{sec: persistent misalignment}. Samples drawn from the high-dimensional PMs of non-robust models are perceptually indistinguishable from high-frequency noise. However, as the PM dimension compresses toward that of the natural data manifold dimension in highly robust models, we observe the emergence of macroscopic semantic structures, such as distinct textures, recognizable object sub-parts, and coherent color clustering. This indicates that dimensional alignment induces perceptual alignment, even in highly complex, high-resolution ambient spaces.

\section{Additional figures}

\begin{figure}[h]
    \centering
    \includegraphics[width=0.9\linewidth]{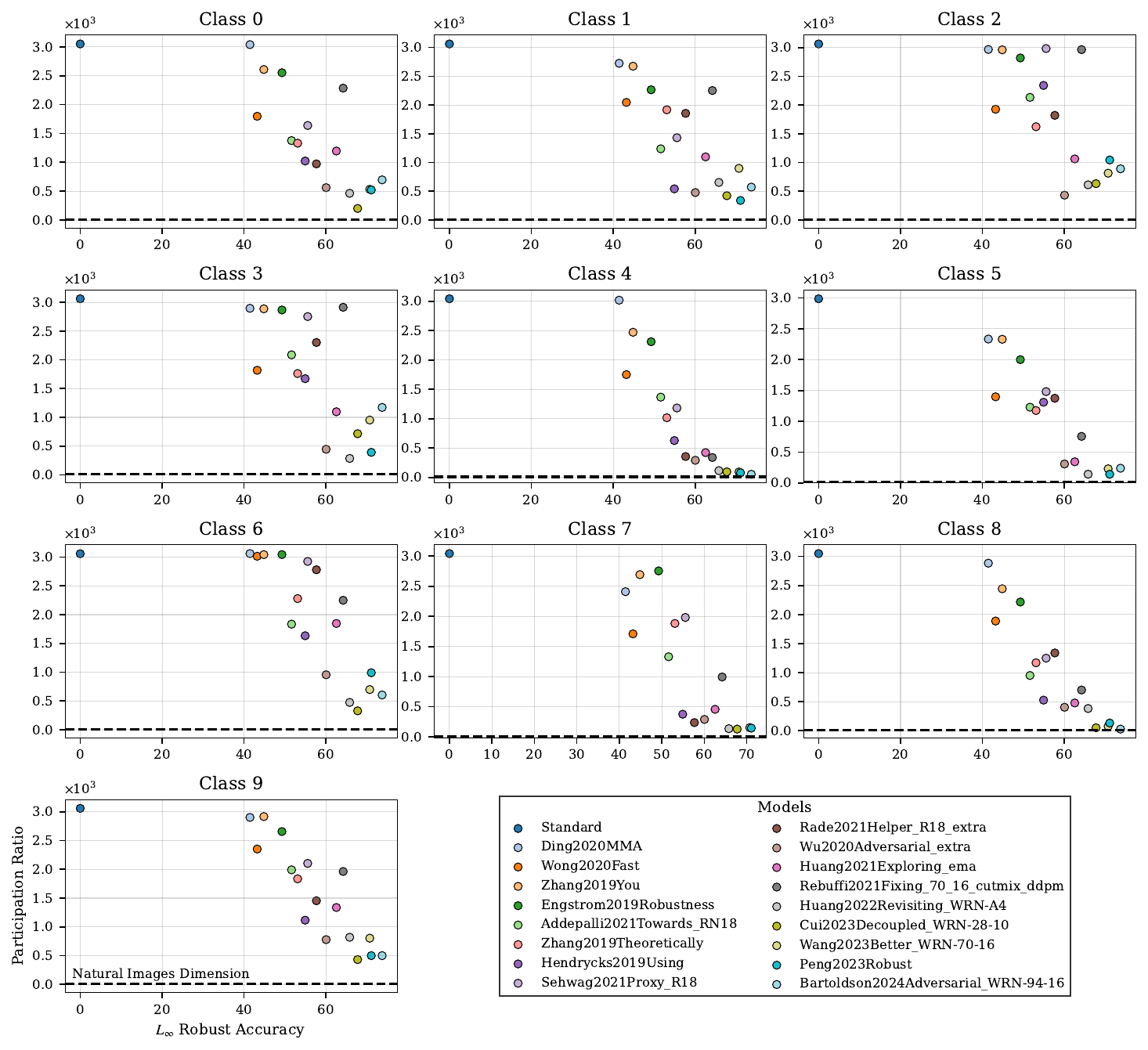}
    \caption{Participation Ratio of the Perceptual Manifold (conditioned on each CIFAR10 class) versus $L_\infty$ robust accuracy for models from RobustBench \citep{croce_robustbench_2021}. The robust accuracy is the one reported on the RobustBench leaderboard.}
    \label{fig: PR vs LinfAcc all}
\end{figure}

\begin{figure}[h]
    \centering
    \includegraphics[width=0.9\linewidth]{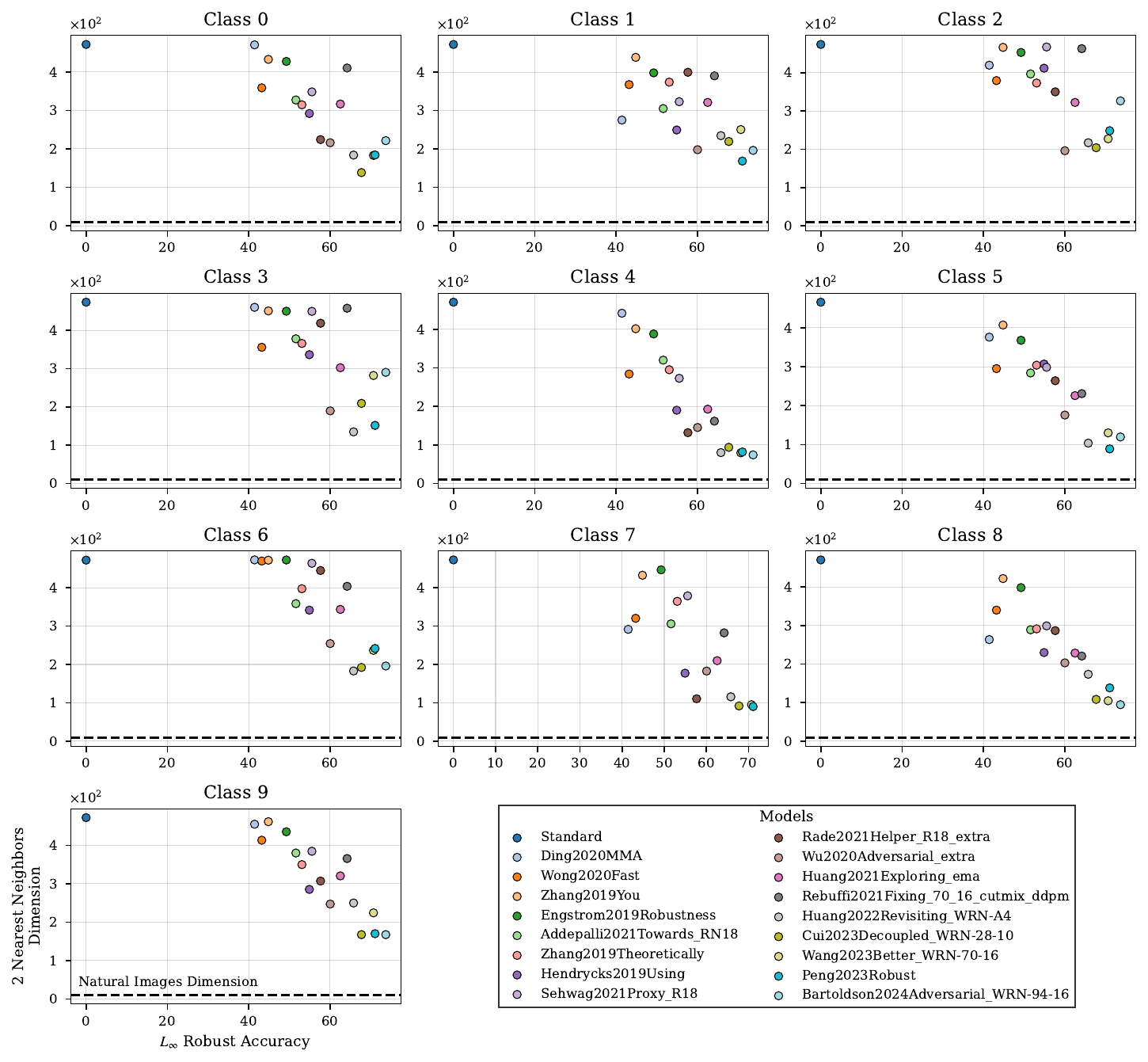}
    \caption{Two Nearest Neighbor dimension of the Perceptual Manifold (conditioned on each CIFAR10 class) versus $L_\infty$ robust accuracy for models from RobustBench \citep{croce_robustbench_2021}. The robust accuracy is the one reported on the RobustBench leaderboard. Note that the values for the 2NN are lower bounds, however, based on the scaling of the predicted intrinsic dimension with dataset size in \cref{fig:scaling 2NN}, we are confident the relative ranking will stay the same}
    \label{fig: 2NN vs Linfacc all}
\end{figure}

\begin{figure}
    \centering
    \includegraphics[width=0.9\linewidth]{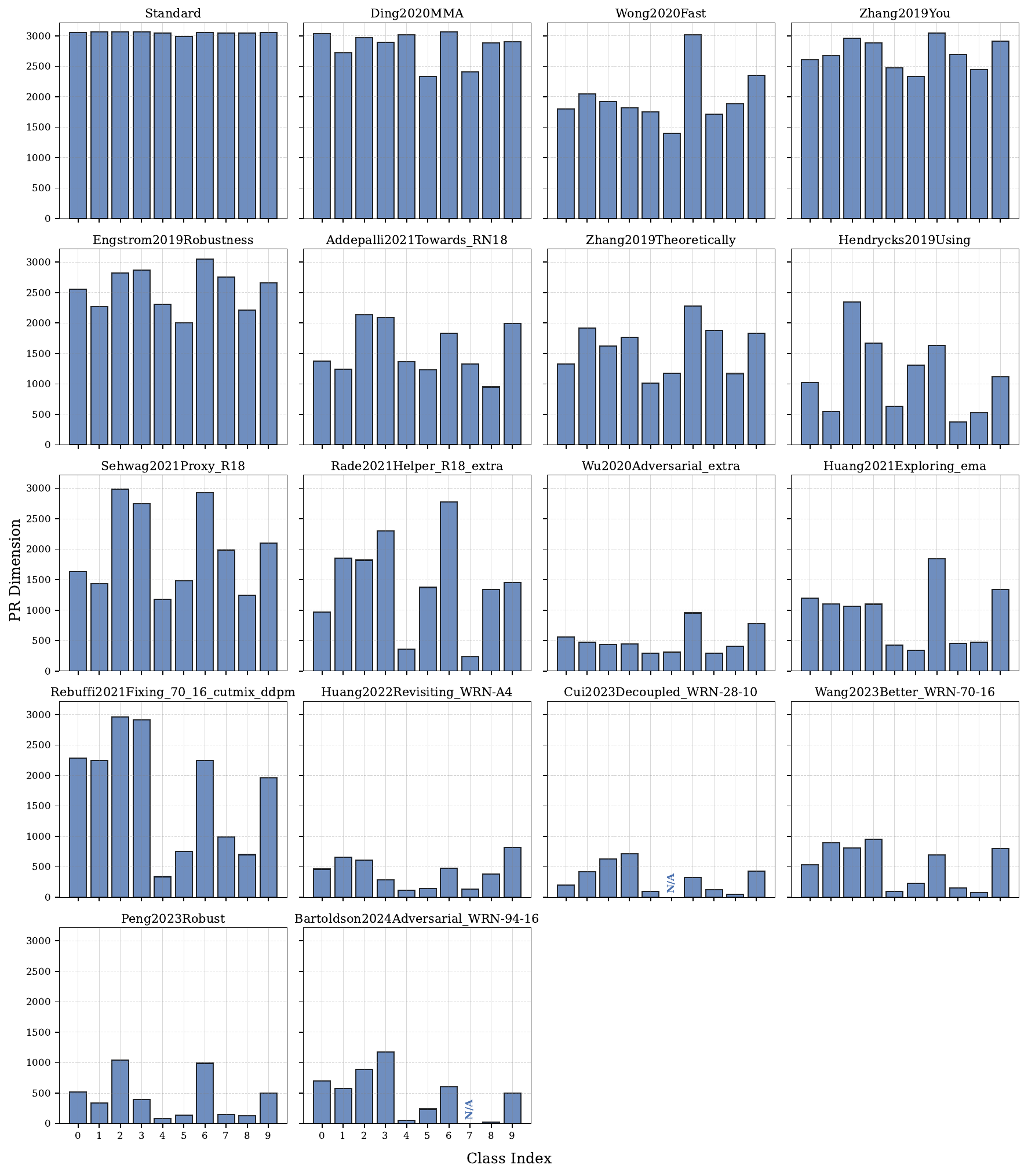}
    \caption{Barplot of the Participation Ratio (PR) for all 10 CIFAR10 classes for each model}
    \label{fig: hetero pr}
\end{figure}

\begin{figure}
    \centering
    \includegraphics[width=0.9\linewidth]{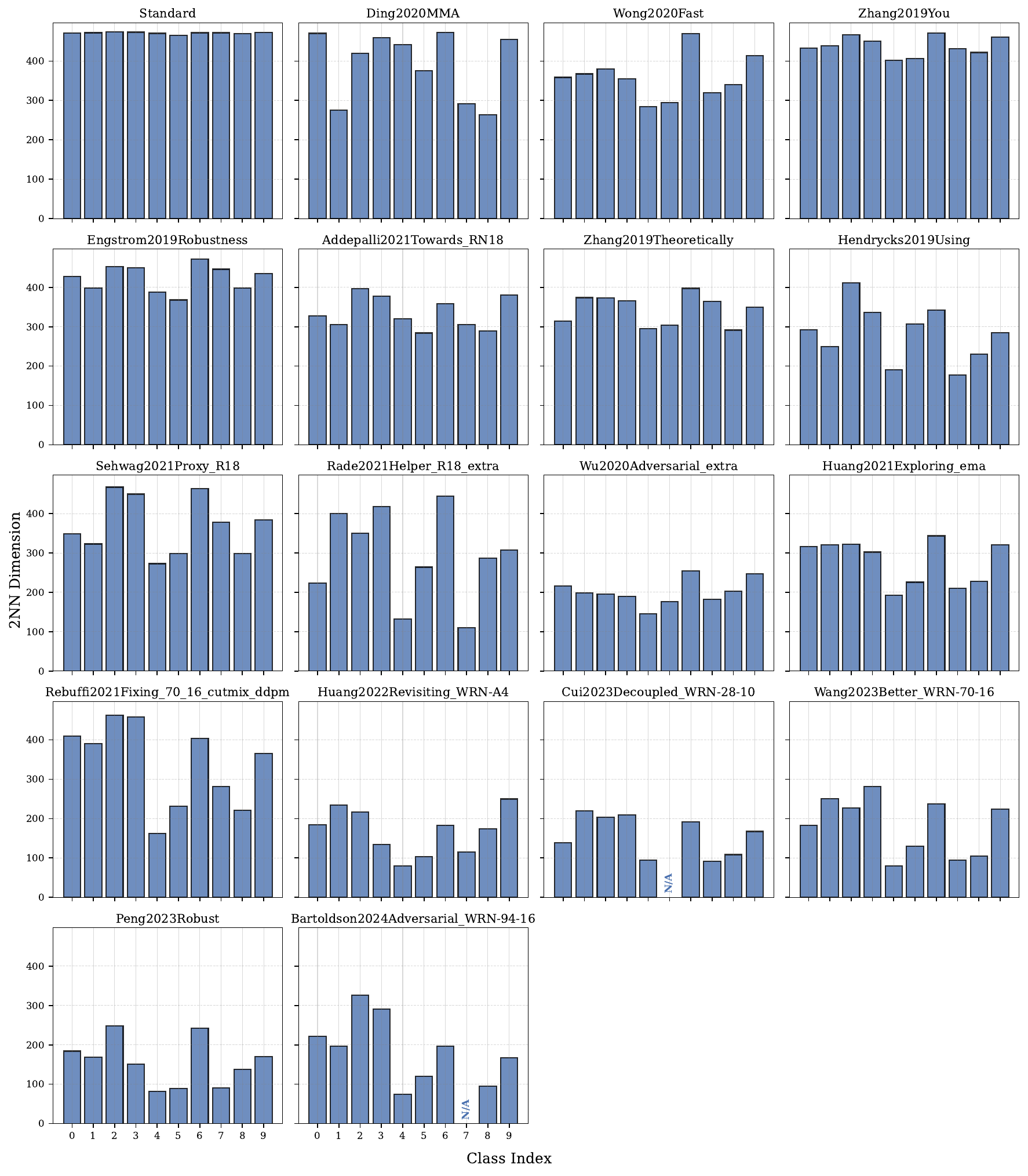}
    \caption{Barplot of the Two Nearest Neighbor (2NN) dimensionality for all 10 CIFAR10 classes for each model}
    \label{fig: hetero 2nn}
\end{figure}

\begin{figure}[h]
    \centering
    \includegraphics[width=0.9\linewidth]{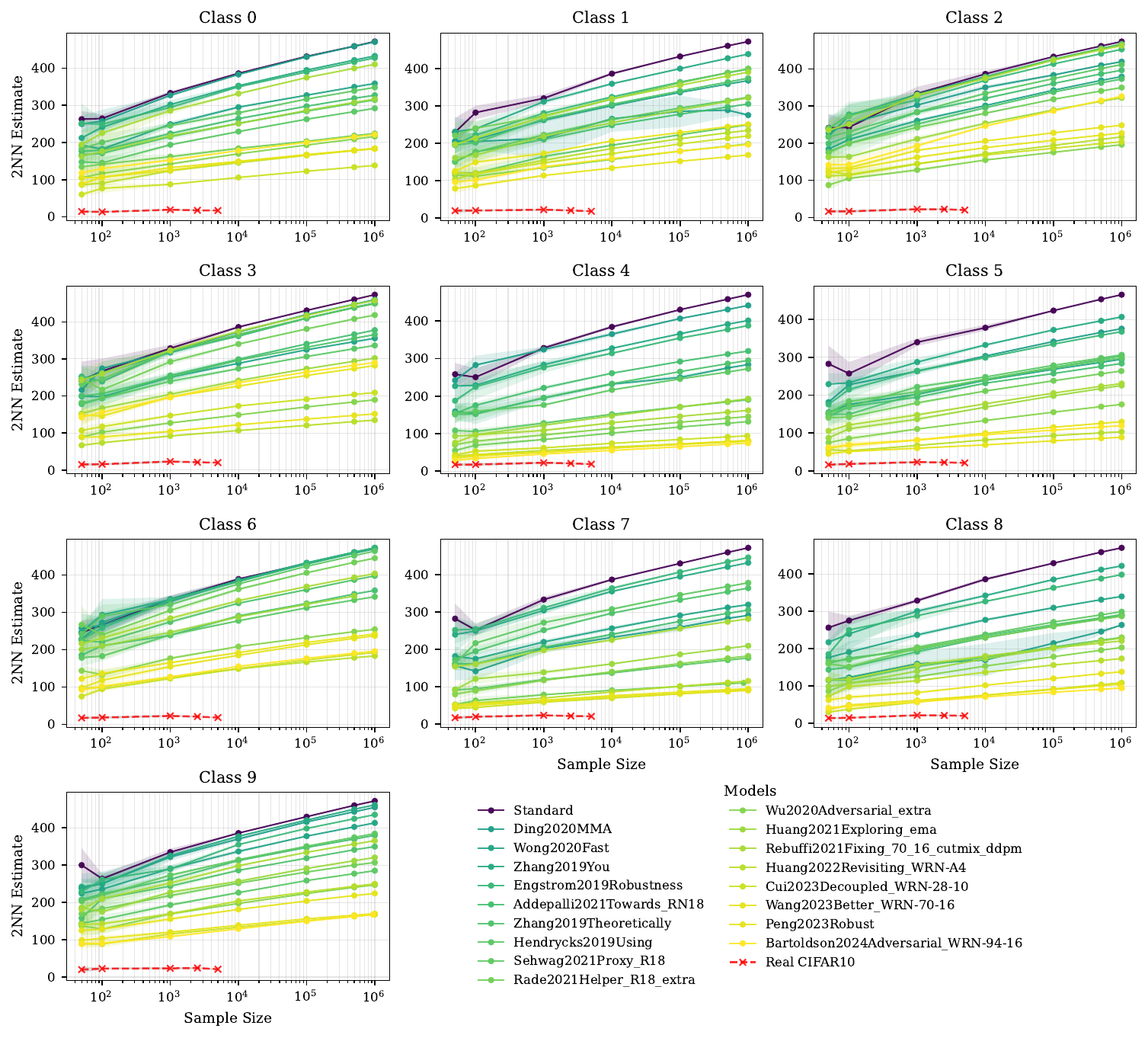}
    \caption{Convergence of Two Nearest Neighbors (2NN) estimates with sample size. We compute 2NN estimates on subsamples of the full dataset ($N=10^6$) to analyze scaling behavior. Due to the curse of dimensionality, a convergence plateau is not observable within the available sample limit; instead, the estimates scale as $\sim\ln N$. However, the stratification of curves—corresponding to models with varying degrees of adversarial robustness—remains consistent across all scales. This suggests that the relative ranking of model manifold dimensionality is invariant to sample size even prior to convergence.}
    \label{fig:scaling 2NN}
\end{figure}

\begin{figure}[h]
    \centering
    \includegraphics[width=0.9\linewidth]{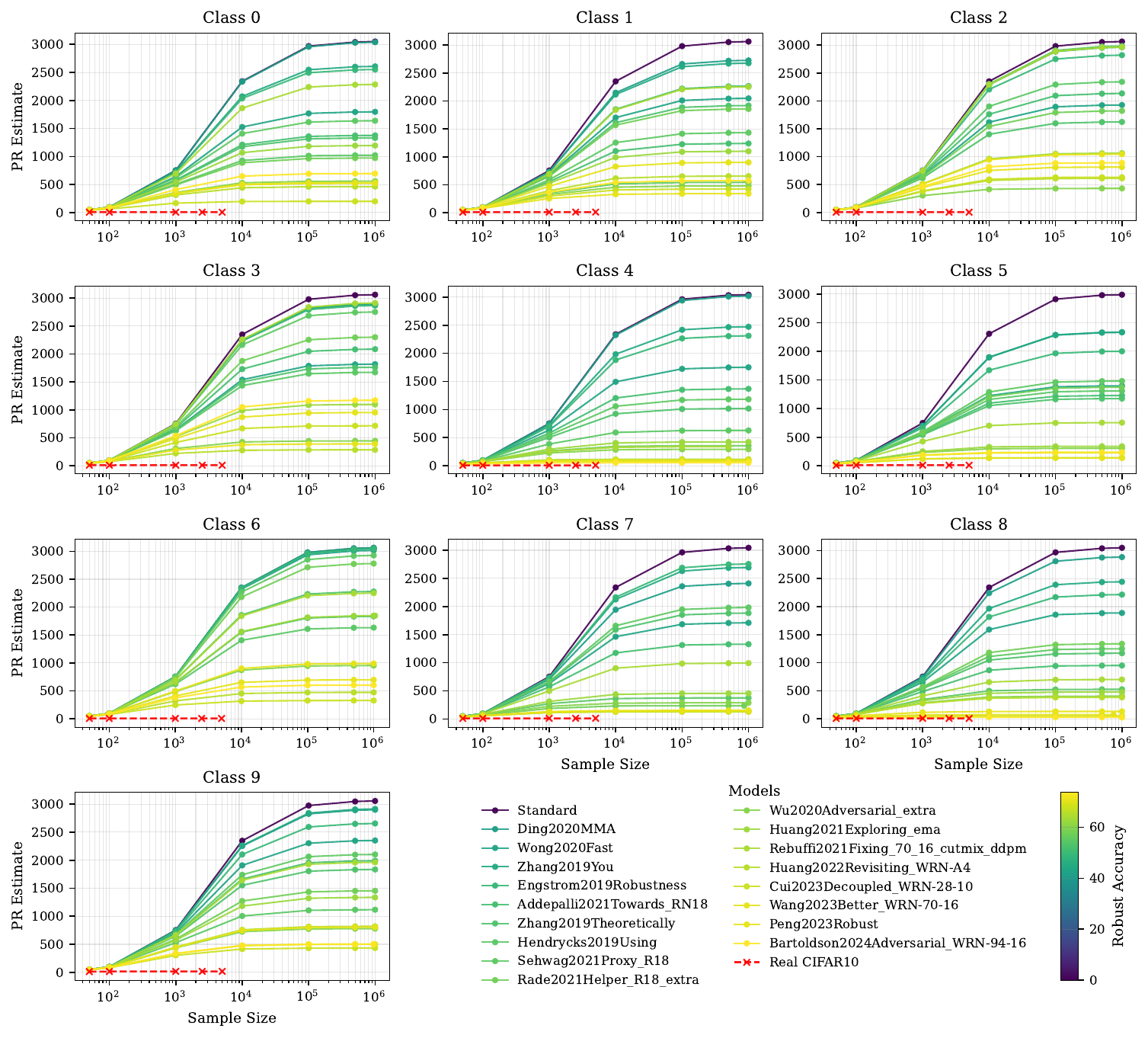}
    \caption{Convergence of Participation Ratio (PR) estimates with sample size. We compute PR estimates on subsamples of the full dataset ($N=10^6$) to analyze scaling behavior. The estimates for all classes plateau as $N$ approaches $10^6$, indicating that the dataset size is sufficient for stable dimensionality estimation. The ordered stratification of curves corresponding to models of increasing robustness shows once again that more robust models yield lower PR values}
    \label{fig:scaling PR}
\end{figure}

\begin{figure}[h]
    \centering
    \includegraphics[width=0.9\linewidth]{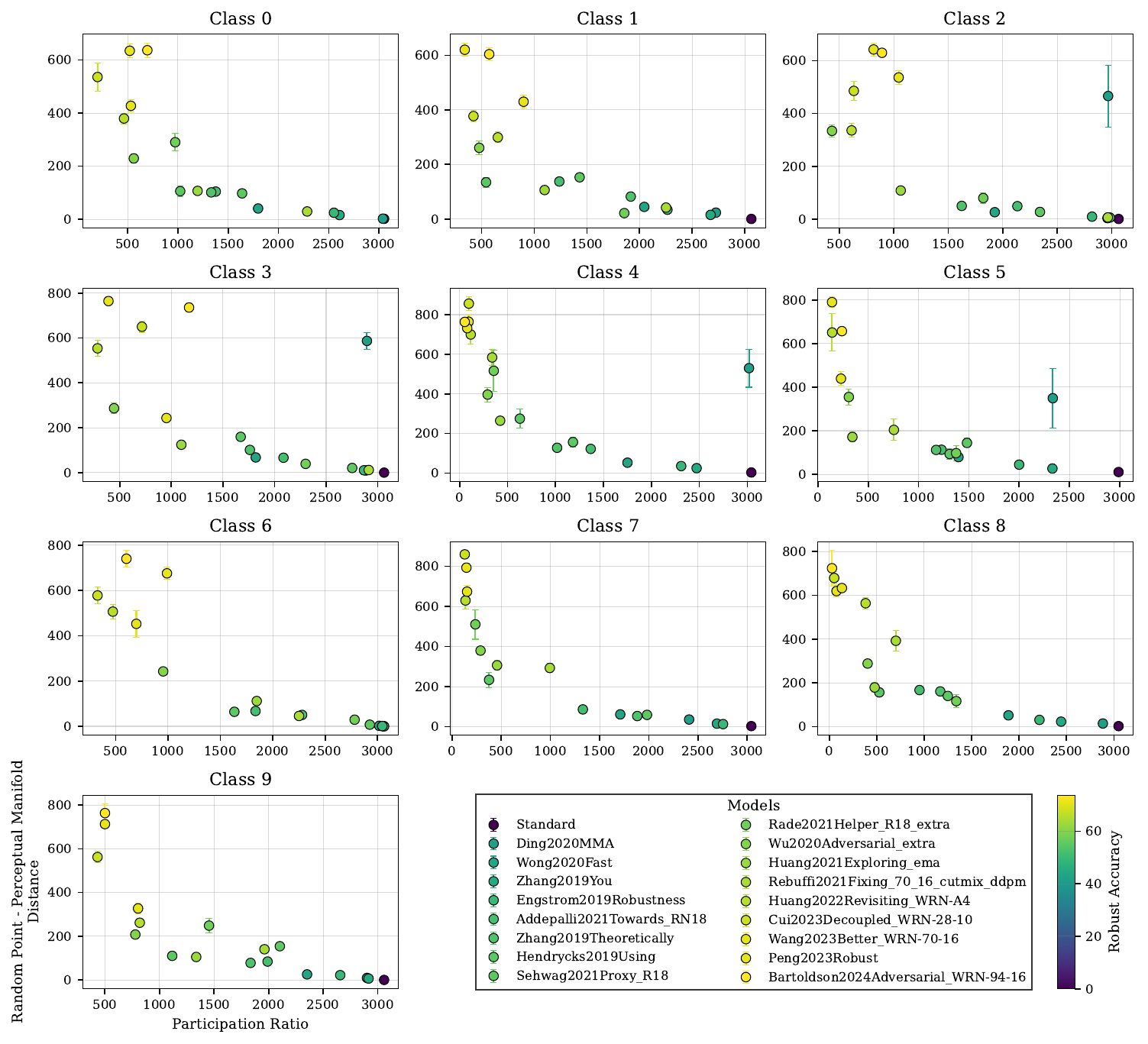}
    \caption{Typical distance between a random point and the class-conditional PM as a function of the Participation Ratio for each class. Error bars represent the $\pm \sigma$ interval.}
    \label{fig:distance vs pr}
\end{figure}

\begin{figure}[h]
    \centering
    \includegraphics[width=0.9\linewidth]{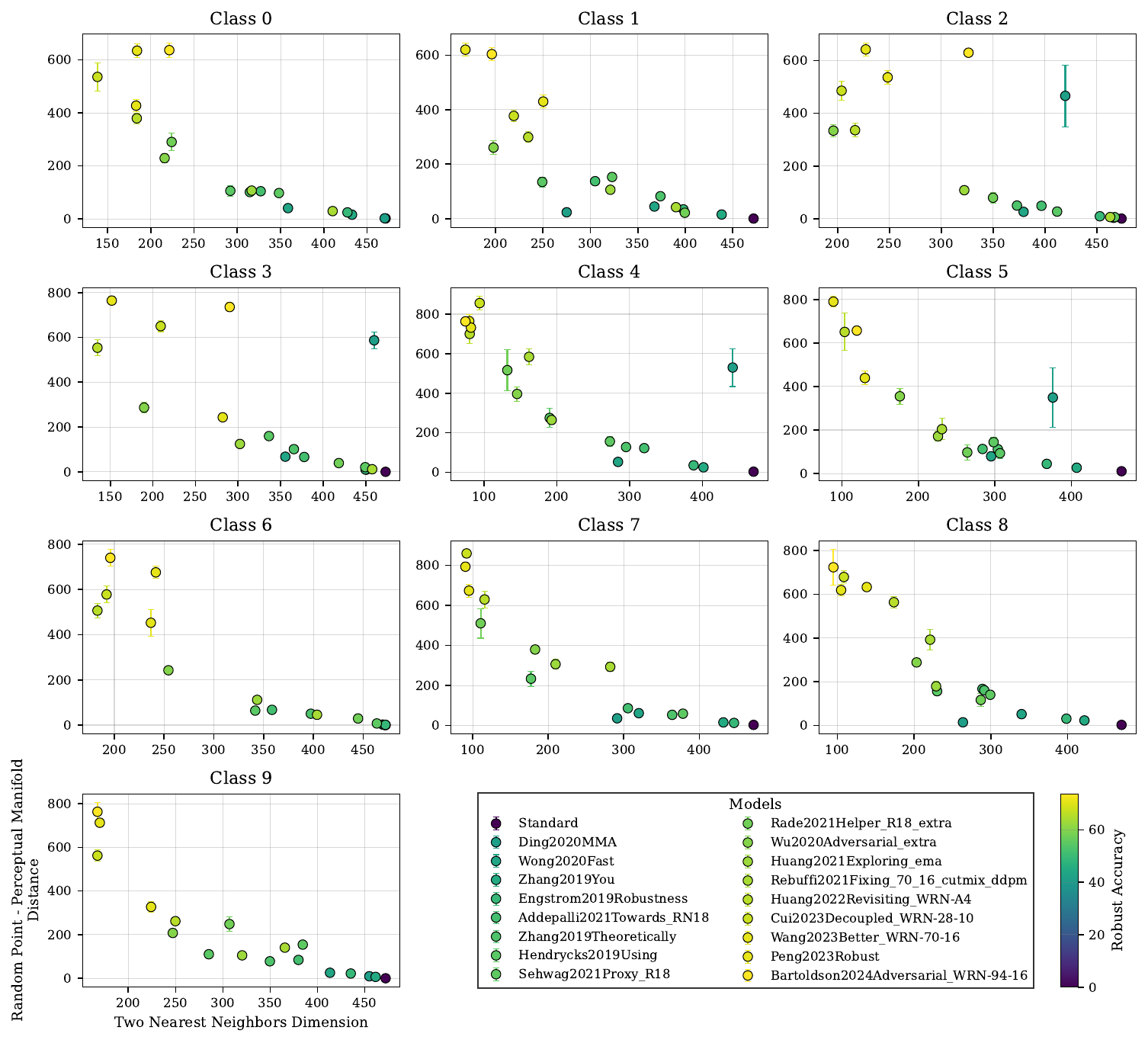}
    \caption{Typical distance between a random point and the class-conditional PM as a function of the Two Nearest Neighbor dimension for each class. Error bars represent the $\pm \sigma$ interval.}    \label{fig:distance vs 2NN}
\end{figure}

\begin{figure}
    \centering
    \includegraphics[width=1\linewidth]{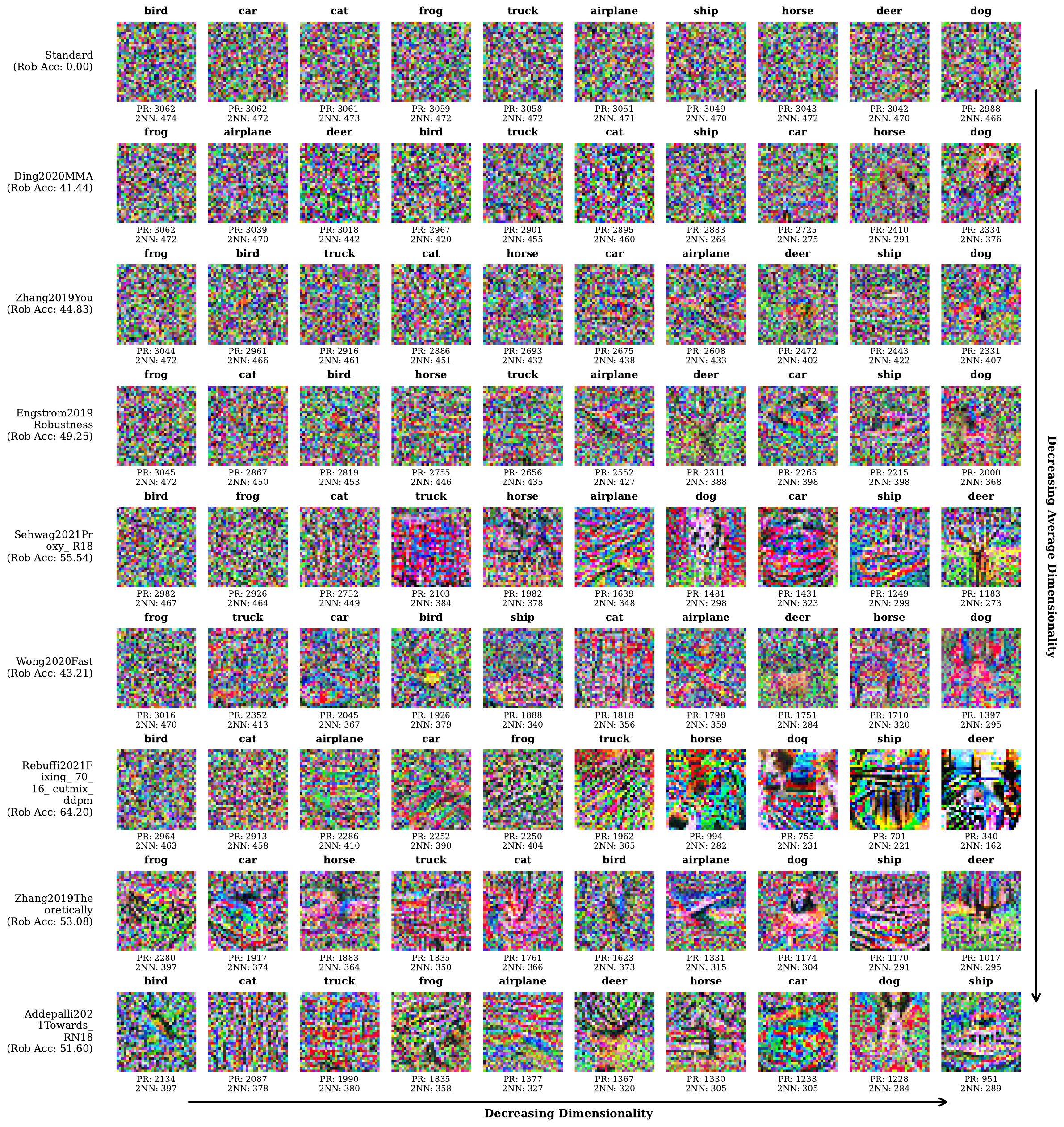}
    \caption{pt 1. Samples from the PM for all CIFAR-10 classes across the 9 models whose PM have the highest average (over classes) dimensionality. Rows are sorted in descending order top to bottom by the model's average PR, which is correlated with the robust accuracy. Columns (classes) are sorted by the PR of the corresponding PM descending from left to right.}
    \label{fig: PM samples pt 1}
\end{figure}
\begin{figure}
    \centering
    \includegraphics[width=1\linewidth]{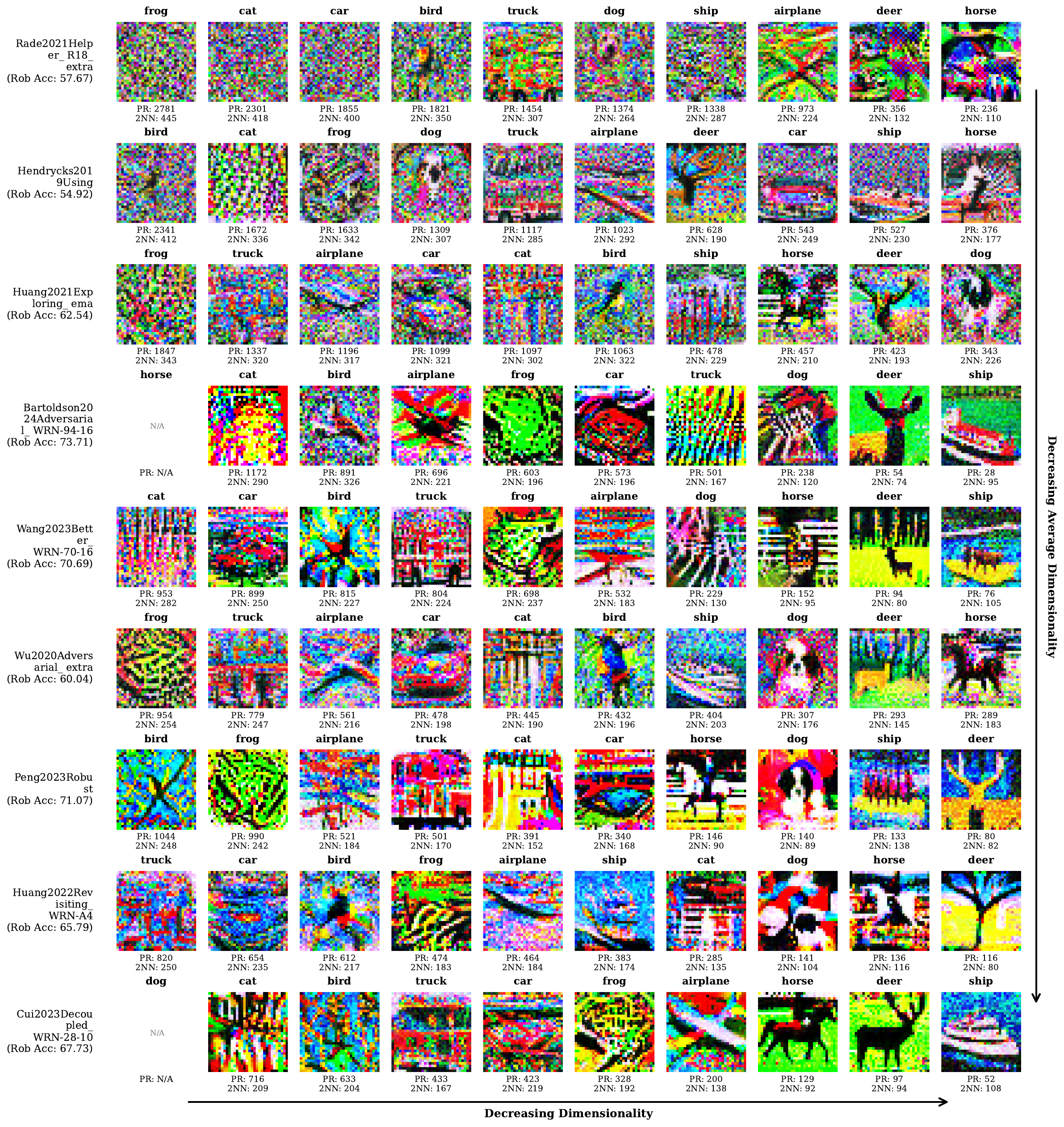}
    \caption{pt 2. Samples from the PM for all CIFAR-10 classes across the 9 models whose PM have the lowest average (over classes) dimensionality. Rows are sorted in descending order top to bottom by the model's average PR, which is correlated with the robust accuracy. Columns (classes) are sorted by the PR of the corresponding PM descending from left to right.}
    \label{fig: PM samples pt2}
\end{figure}

\begin{figure}
    \centering
    \includegraphics[width=0.9\linewidth]{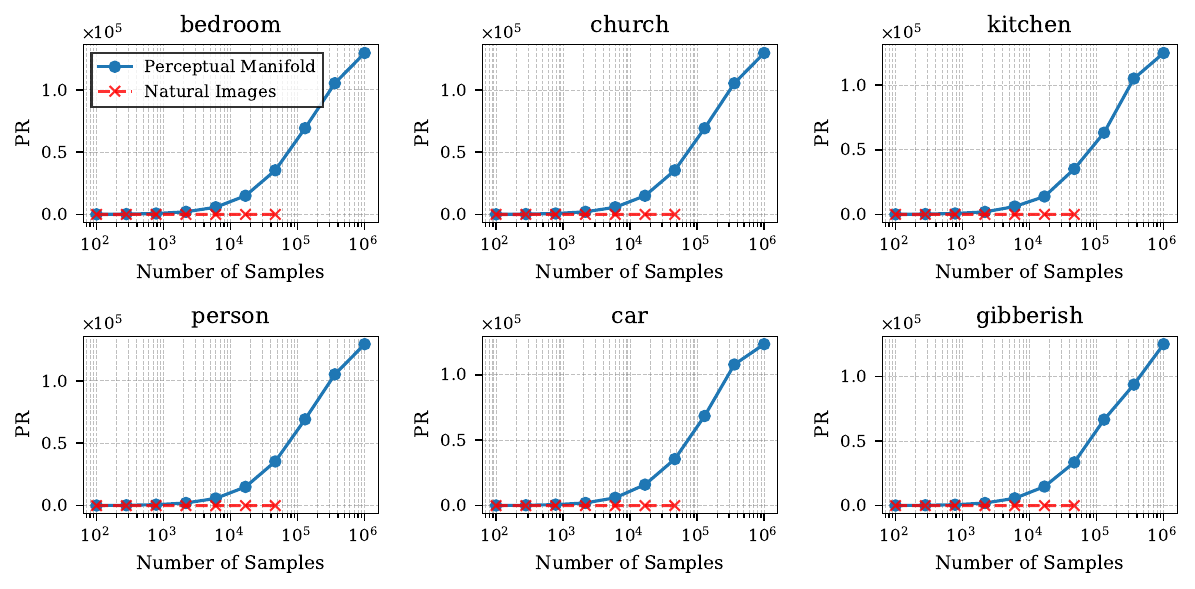}
    \caption{How the computed Participation Ratio scales with dataset size. $10^6$ samples are actually not enough for the estimation of the PM's dimension to converge, so value needs to be interpreted as lower values. That of natural images, on the other hand, has converged}
    \label{fig: scaling clip pr}
\end{figure}

\begin{figure}
    \centering
    \includegraphics[width=0.9\linewidth]{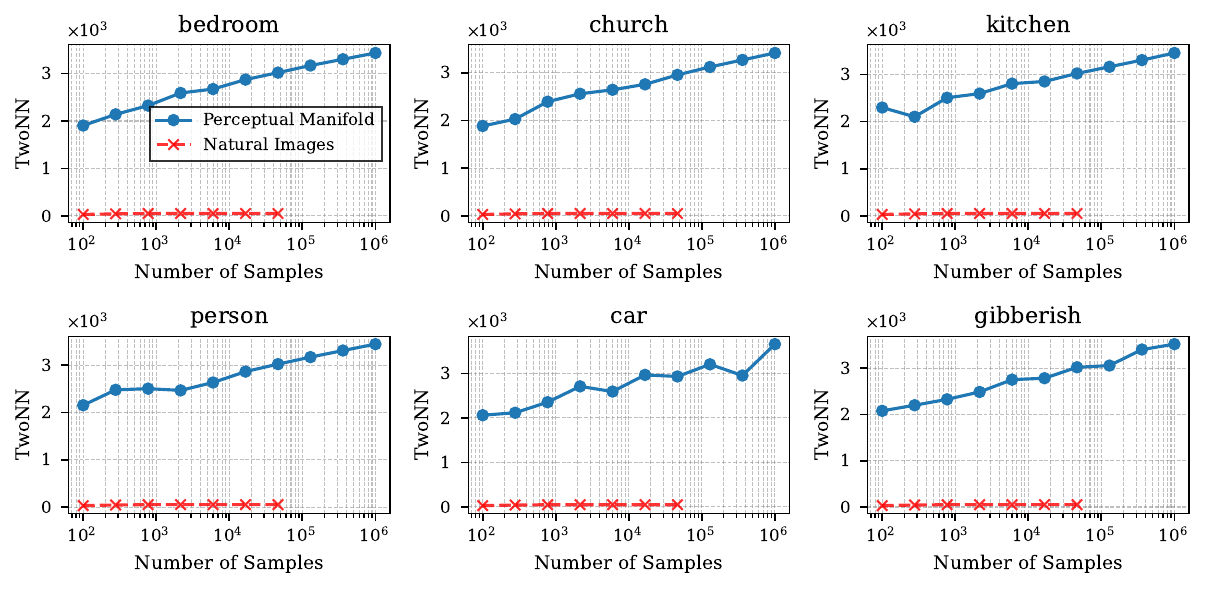}
    \caption{How the computed Two Nearest Neighbor dimension scales with dataset size. $10^6$ samples are actually not enough for the estimation of the PM's dimension to converge, so value needs to be interpreted as lower values.  That of natural images, on the other hand, has converged}
    \label{fig: scaling clip two nn}
\end{figure}

\begin{figure}
    \centering
    \includegraphics[width=0.9\linewidth]{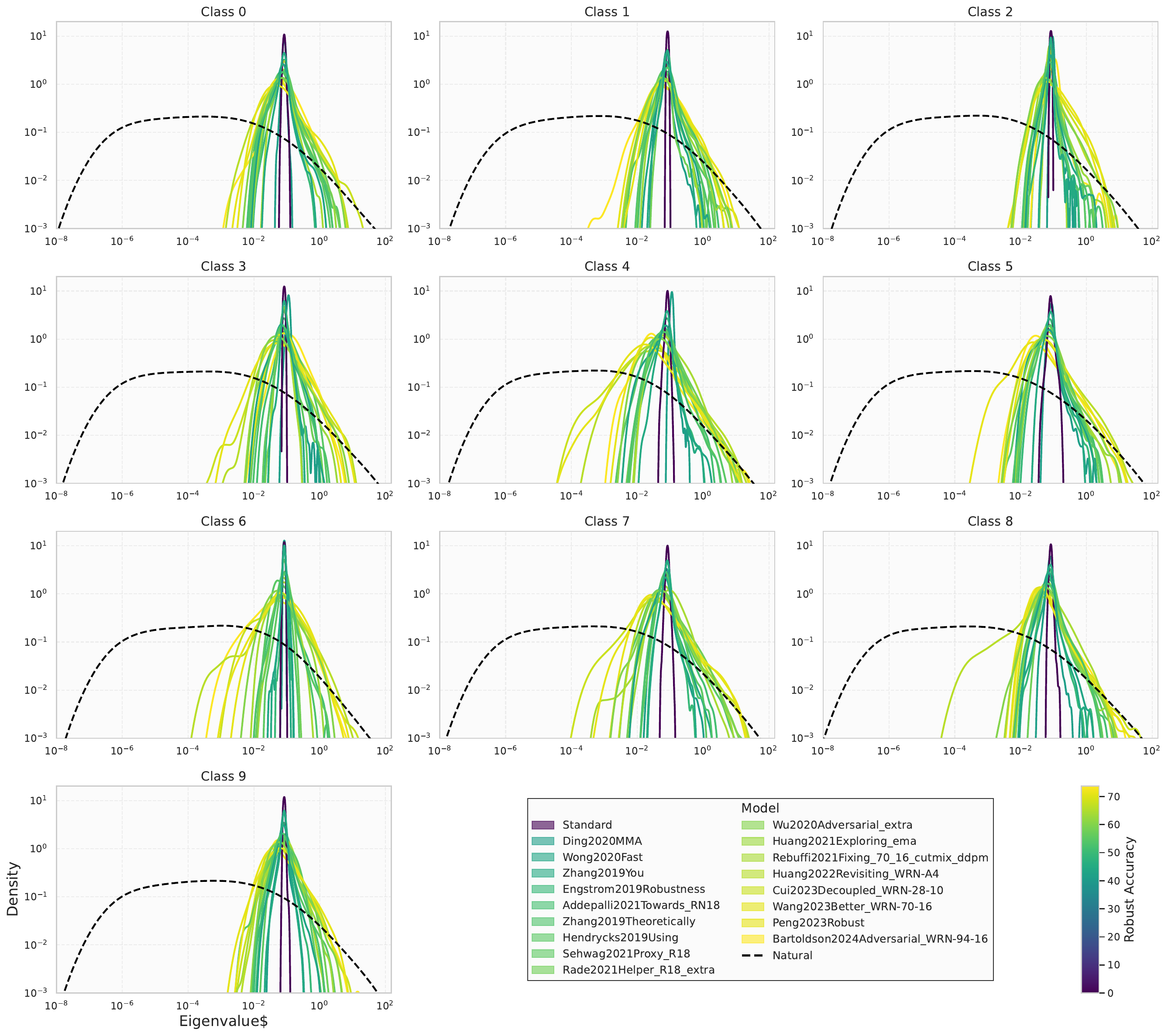}
    \caption{Full colored lines: density of the spectrum of the covariance matrix of the PM of different model. Dashed black line: density of the spectrum of the covariance of natural images. }
    \label{fig: eigenvalue distribution all classes}
\end{figure}

\begin{figure}
    \centering
    \includegraphics[width=0.9\linewidth]{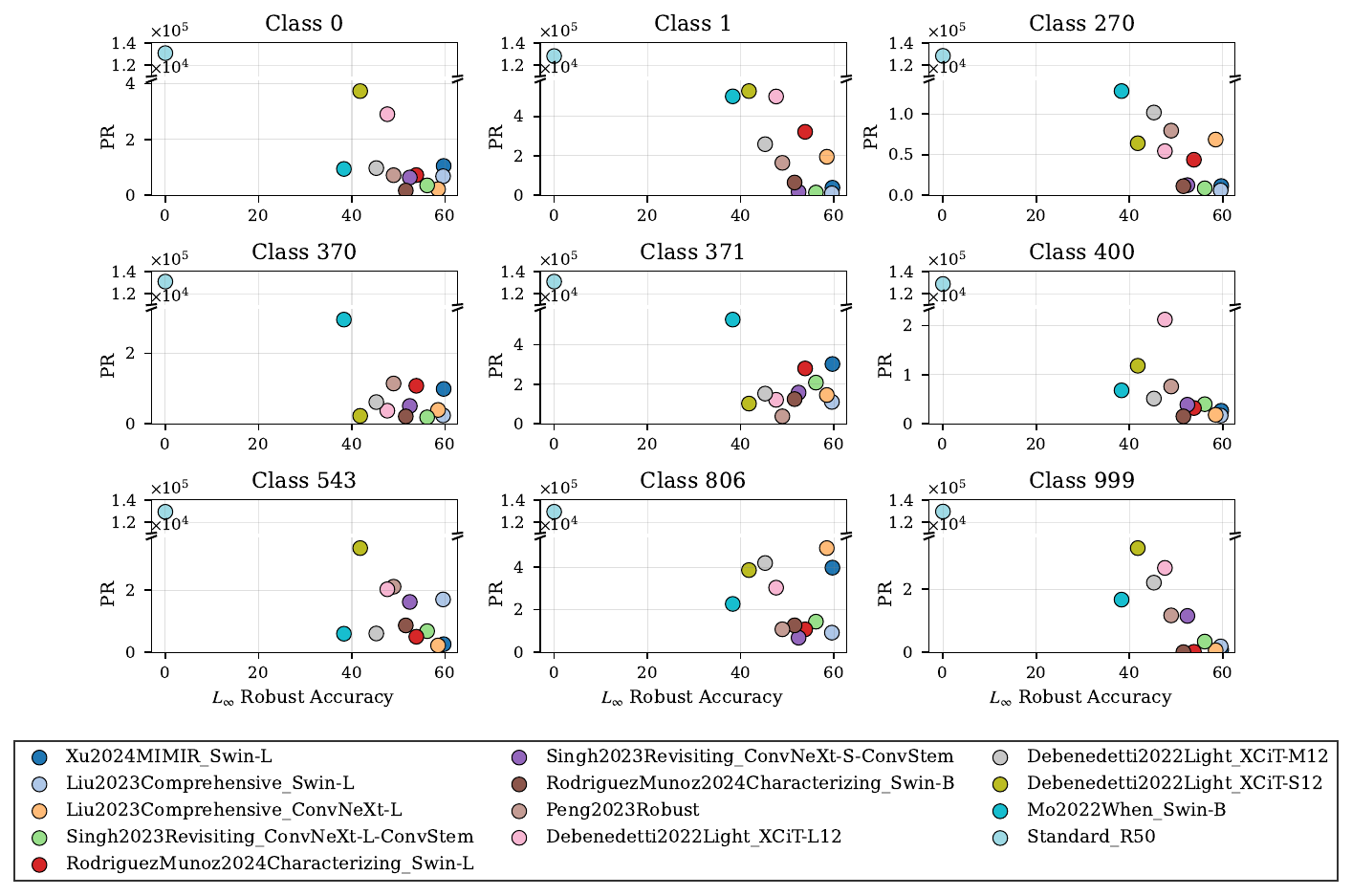}
    \caption{Participation Ratio of the Perceptual Manifold (each subplot conditioned on a different ImageNet class) versus $L_\infty$ robust accuracy (as reported in RobustBench \citep{croce_robustbench_2021})}
    \label{fig:PR_vs_robacc_by_class_imagenet}
\end{figure}

\begin{figure}
    \centering
    \includegraphics[width=0.9\linewidth]{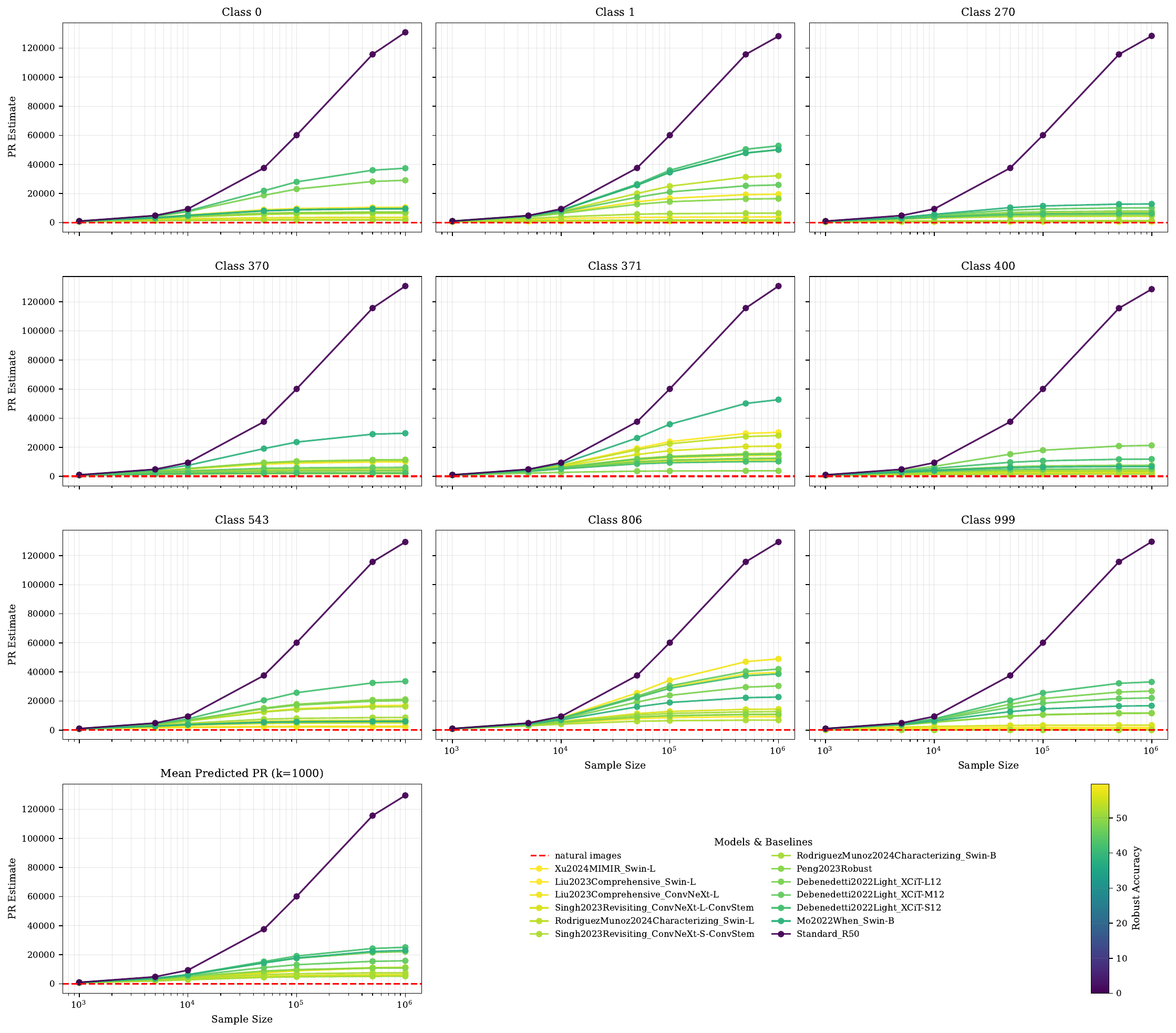}
    \caption{Convergence of Participation Ratio (PR) estimates with sample size. We compute PR estimates on subsamples of the full dataset ($N=10^6$) to analyze scaling behavior. The estimates for all robust models plateau as N approaches $10^6$, while that for the standard model doesn't and is therefore to be taken as a lower bound. The ordered stratification of curves corresponding to models of increasing robustness shows once again that more robust models yield lower PR values}
    \label{fig:PR_robac_scaling_imagenet}
\end{figure}

\begin{figure}
    \centering
    \includegraphics[width=0.9\linewidth]{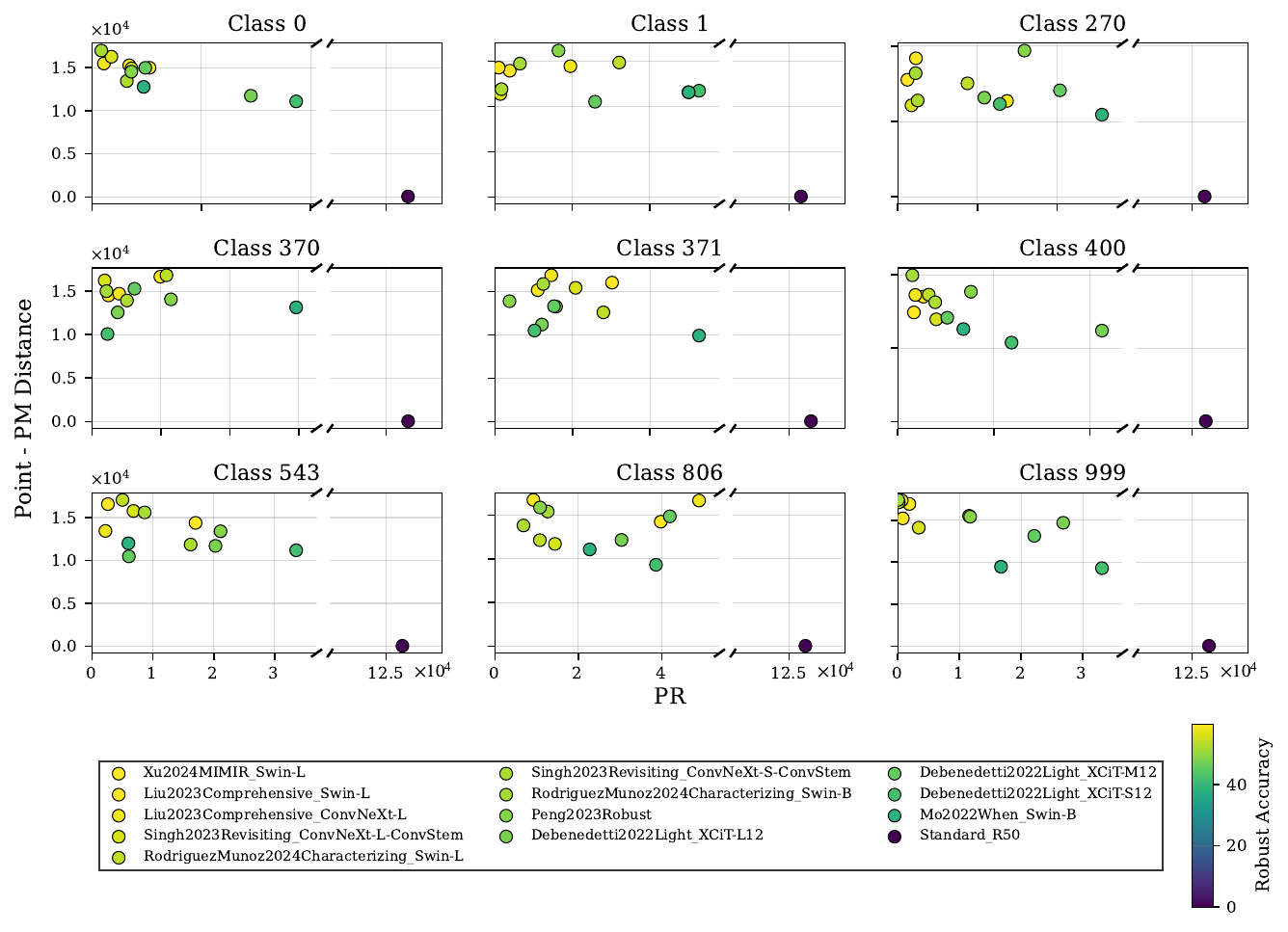}
    \caption{Typical distance between a random point and the class-conditional PM as a function of the Participation Ratio for 9 ImageNet Classes. The color scheme report the robust accuracy}
    \label{fig:distance_vs_pr_all_classes_imagenet}
\end{figure}


\end{document}